\documentclass{article} % For LaTeX2e
\usepackage{iclr2025_conference,times}

\usepackage{amsmath,amsfonts,bm}

\def\eqref#1{equation~\ref{#1}}
\def\1{\bm{1}}

\def\vc{{\bm{c}}}

\def\vv{{\bm{v}}}

\def\vx{{\bm{x}}}
\def\vy{{\bm{y}}}
\def\vz{{\bm{z}}}

\DeclareMathAlphabet{\mathsfit}{\encodingdefault}{\sfdefault}{m}{sl}
\SetMathAlphabet{\mathsfit}{bold}{\encodingdefault}{\sfdefault}{bx}{n}

\newcommand*\modelname{DART}

\def\veps{{\bm{\epsilon}}}

\usepackage{comment}
\usepackage[utf8]{inputenc} % allow utf-8 input
\usepackage[T1]{fontenc}    % use 8-bit T1 fonts
\usepackage{url}            % simple URL typesetting
\usepackage{booktabs}       % professional-quality tables
\usepackage{amsfonts}       % blackboard math symbols
\usepackage{nicefrac}       % compact symbols for 1/2, etc.
\usepackage{microtype}      % microtypography
\usepackage{mathtools}
\usepackage{wrapfig}
\usepackage{bm}
\usepackage{float}
\usepackage{amsthm}
\usepackage{amsmath}
\usepackage{enumitem}
\usepackage{amssymb}
\usepackage{booktabs}
\usepackage{graphicx}
\usepackage{graphbox}
\usepackage{notes2bib}
\usepackage{multirow}
\usepackage{algorithm}
\usepackage{algpseudocode}
\usepackage{setspace}
\usepackage{cancel}

\definecolor{mydarkblue}{RGB}{0, 2, 115}
\usepackage[colorlinks=true,linkcolor=mydarkblue,citecolor=mydarkblue,urlcolor=mydarkblue,pdfborder={0 0 0}]{hyperref}

\usepackage[capitalize,noabbrev]{cleveref}
\crefname{section}{§}{§§}
\Crefname{section}{§}{§§}
\crefname{theorem}{Theorem}{Theorems}
\crefname{lemma}{Lemma}{Lemmas}
\crefname{equation}{Eq}{Eqs}
\crefname{proposition}{Proposition}{Propositions}
\crefname{claim}{Claim}{Claims}
\crefname{appendix}{Appendix}{Appendices}
\crefname{algorithm}{Algorithm}{Algorithms}
\crefname{figure}{Figure}{Figures}
\crefname{table}{Table}{Tables}
\crefname{remark}{Remark}{Remarks}
\crefname{definition}{Def.}{Definitions}
\crefname{corollary}{Corollary}{Corollaries}

\usepackage{tcolorbox}
\usepackage{lipsum} % For dummy text, can be removed

\newtheoremstyle{fancy}
  {10pt} % Space above
  {10pt} % Space below
  {\itshape} % Body font
  {} % Indent amount
  {\bfseries} % Theorem head font
  {.} % Punctuation after theorem head
  {5pt plus 1pt minus 1pt} % Space after theorem head
  {\thmname{#1}\thmnumber{ #2}\thmnote{ (#3)}} % Theorem head spec

\theoremstyle{fancy}
\newtheorem{proposition}{Proposition}

\title{\modelname{}: Denoising Autoregressive Transformer for Scalable Text-to-Image Generation}

\iclrfinalcopy

\author{Jiatao Gu$^\dagger$,  Yuyang Wang$^\dagger$, Yizhe Zhang$^\dagger$, Qihang Zhang$^\delta$\thanks{Work done as part of an internship at Apple.}, 
\\ 
\textbf{Dinghuai Zhang}$^{\gamma*}$,
\textbf{Navdeep Jaitly$^\dagger$, Josh Susskind$^\dagger$, Shuangfei Zhai$^\dagger$} \\
$^\dagger$Apple, $^\delta$The Chinese University of Hong Kong, $^\gamma$Mila\\
\texttt{$^\dagger$\{jgu32,yuyangw,yizzhang,njaitly,jsusskind,szhai\}@apple.com}\\
\texttt{$^\delta$qhzhang@link.cuhk.edu.hk} \ \texttt{$^\gamma$dinghuai.zhang@mila.quebec}
}

\iclrfinalcopy % Uncomment for camera-ready version, but NOT for submission.
\begin{document}

\maketitle

\begin{figure}[h]
    \centering
    \includegraphics[width=0.945\textwidth]{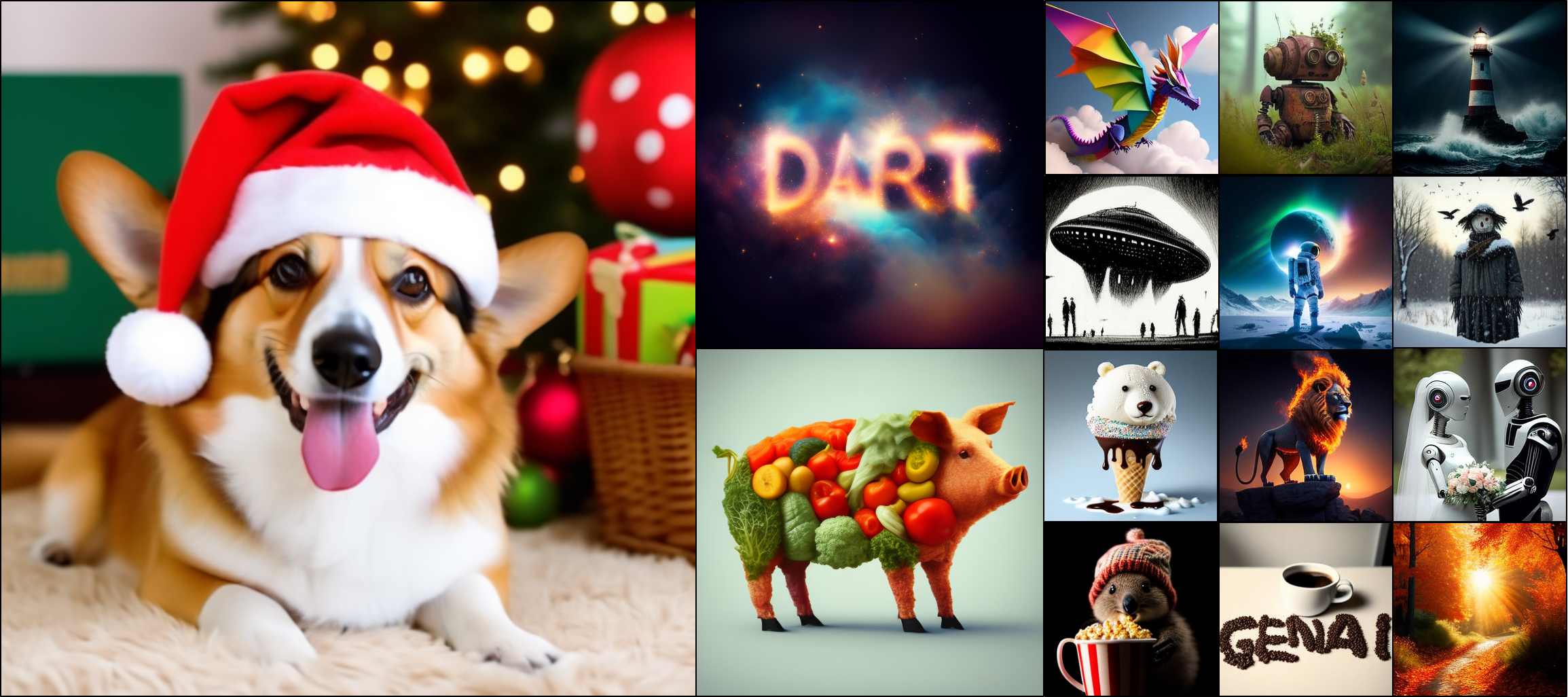}
    \caption{Curated examples of images generated by DART at $256^2$, $512^2$ and $1024^2$ pixels.} %\yz{Shall we 1) use different colors to differentiate a and b, and 2) explicitly say the generation from b is more diverse in style? }\jg{Sounds good}}
    \label{fig:teaser}
\end{figure}
\begin{abstract}
Diffusion models have become the dominant approach for visual generation. They are trained by denoising a Markovian process which gradually adds noise to the input. We argue that the Markovian property limits the model’s ability to fully utilize the generation trajectory, leading to inefficiencies during training and inference. In this paper, we propose DART, a transformer-based model that unifies autoregressive (AR) and diffusion within a non-Markovian framework.  DART iteratively denoises image patches spatially and spectrally using an AR model that has the same architecture as standard language models. DART does not rely on image quantization, which enables more effective image modeling while maintaining flexibility. Furthermore, DART seamlessly trains with both text and image data in a unified model. Our approach demonstrates competitive performance on class-conditioned and text-to-image generation tasks, offering a scalable, efficient alternative to traditional diffusion models. Through this unified framework, DART sets a new benchmark for scalable, high-quality image synthesis.
\end{abstract}
\section{Introduction}
\begin{figure}[t]
    \centering
    \includegraphics[width=0.965\textwidth]{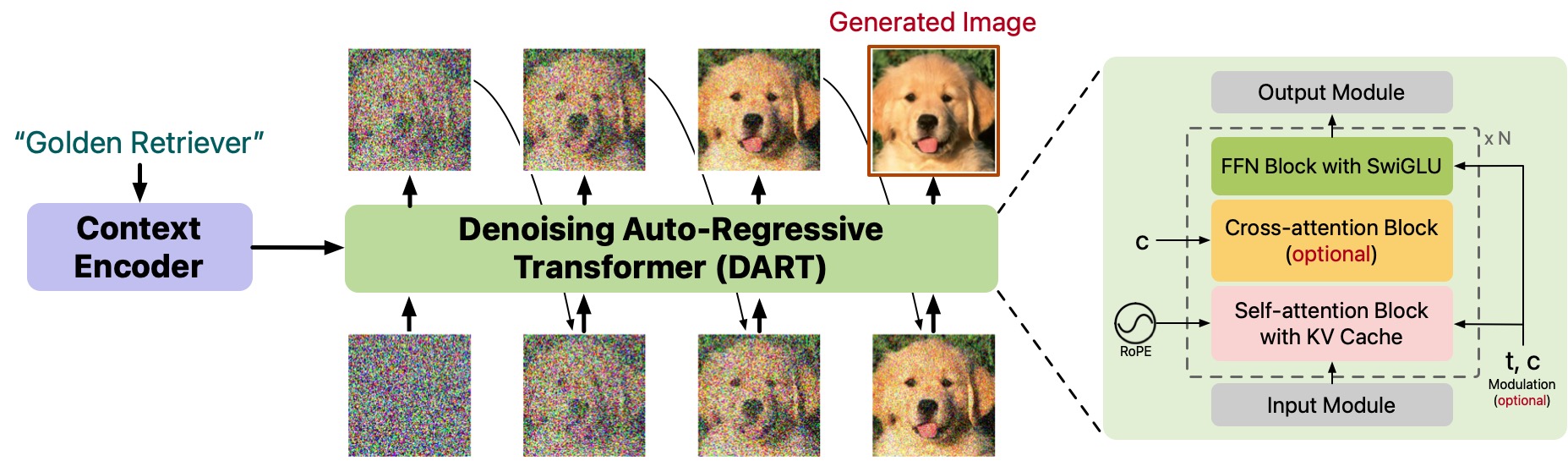}
    \caption{($\leftarrow$) A general illustration of the proposed \modelname{}. The model autoregressively denoises image through a Transformer until a clean image is generated. Here, whole images are shown for visualization purpose; %\protect\footnotemark. 
    ($\rightarrow$) We show the architecture details which integrates state-of-the-art designs similar to common language models~\citep{dubey2024llama}.
    }%It can also generate text through next token prediction. Both images and texts are generated in an autoregressive manner. Conditions like text prompt are also added through a cross-attention.}
    \vspace{-10pt}
    \label{fig:dart_pipeline}
\end{figure}
Recent advancements in deep generative models have led to significant breakthroughs in visual synthesis, with diffusion models emerging as the dominant approach for generating high-quality images~\citep{rombach2022high,esser2024scaling}. Diffusion models~\citep{sohl2015deep, ho2020denoising} operate by progressively adding Gaussian noise to an image and learning to reverse this process in a sequence of denoising steps. 
Despite their success, these models are difficult to train on high resolution images directly, requiring either cascaded models~\citep{ho2022cascaded}, or multiscale approaches~\citep{gu2023matryoshka} or preprocessing of images to autoencoder codes at lower resolutions~\citep{rombach2022high}. These limitations can stem from their reliance on the Markovian assumption, which simplifies the generative process but restricts the model only to see the generation from the previous step. 
This often leads to inefficiencies during training and inference, as the individual steps of denoising are unaware of the trajectory of generations from prior steps.
%as each step is conditioned only on the previous one, discarding potentially valuable information from earlier stages. 
%Several works have aimed to address these limitations by introducing techniques that improve the efficiency of diffusion models during sampling. For example, Denoising Diffusion Implicit Models (DDIMs)~\cite{song2020denoising} introduce a non-Markovian formulation for the reverse denoising process, allowing the model to generate images in fewer steps without significantly compromising image quality. However, the diffusion model still follows Markovian assumption during training, which discards informative diffusion trajectory. 
%\footnotetext{We here show input as whole images for visualization purpose} %. In practice, \modelname{} takes sequence of tokens as input and denoises them autoregressively. See \S \ref{sec:model} and \cref{fig:attn_mask} for more details. }

In parallel, autoregressive models, such as GPT-4~\citep{achiam2023gpt}, have shown great success in modeling long-range dependencies in sequential data, particularly in the field of natural language processing. 
These models efficiently cache computations and manage dependencies across time steps, which has also inspired research into adapting autoregressive models for image generation. Early efforts such as PixelCNN~\citep{Oord2016} suffered from high computational costs due to pixel-wise generation. More recent models like VQ-GAN~\citep{esser2021taming} and related work~\citep{yu2022scaling,team2024chameleon,tian2024visual} learn models of quantized images in a compressed latent space; 
%However, these approaches still rely on image quantization, which introduces its own set of challenges.
\citet{li2024autoregressive} propose to generate directly in such space without quantization by employing a diffusion-based loss function.
However, these methods fail to fully leverage the progressive denoising benefits of diffusion models, resulting in limited global context and error propagation during generation.

To address these limitations, we propose \underline{D}enoising \underline{A}uto\underline{R}egressive \underline{T}ransformer (\modelname{}), a novel generative model that integrates autoregressive modeling within a non-Markovian diffusion framework~\citep{song2021denoising} (Fig.~\ref{fig:dart_pipeline}). 
The non-Markovian formulation in \modelname{} enables the model to leverage the full generative trajectory during training and inference, while retaining the progressive modeling benefits of diffusion models, resulting in more efficient and flexible generation compared to traditional diffusion and autoregressive approaches.
Additionally, \modelname{} introduces two key improvements to address the limitations of the non-Markovian approach: (1) token-level autoregressive modeling (\modelname{}-AR), which captures dependencies between image tokens autoregressively, enabling finer control and improved generation quality, and (2) a flow-based refinement module (\modelname{}-FM), which enhances the model’s expressiveness and smooths transitions between denoising steps. These extensions make \modelname{} a flexible and efficient framework capable of handling a wide range of tasks, including class conditional, text-to-image, as well as multimodal generation.

\modelname{} offers a scalable, efficient alternative to traditional diffusion models, achieving competitive performance on standard benchmarks for class-conditioned (e.g., ImageNet~\citep{Deng2009ImageNet:Database}) and text-to-image generation. % This unified framework sets a new benchmark for high-quality image synthesis. and demonstrates the potential of combining the strengths of diffusion and autoregressive approaches.
% where each denoising step depends only on the previous one.
% To address these challenges and unify the strengths of both diffusion and autoregressive models, we propose \modelname{} — a novel approach that leverages autoregressive modeling within a non-Markovian framework for scalable image generation. \modelname{} recursively denoises image patches both spatially and spectrally. Such a framework leverages the diffusion trajectory during training which allows more efficient training. Unlike traditional AR image generation models, \modelname{} operates without the need for image quantization, enabling more effective and flexible image modeling while preserving the continuous nature of the data. This design eliminates many of the inefficiencies associated with sequential pixel-by-pixel generation and avoids the complexity of operating in a quantized latent space. On standard image generation benchmarks including ImageNet~\cite{Deng2009ImageNet:Database} and CC12M~\cite{changpinyo2021cc12m}, \modelname{} demonstrates rival performance with competitive baselines. Furthermore, \modelname{} extends the autoregressive paradigm to text-to-image generation, unifying the training process for both image and text data in a single model. This seamless integration enables \modelname{} to handle a wide variety of generative tasks with a high degree of control and flexibility. 
To summarize, major contributions of our work include: 
\begin{itemize}[leftmargin=*]
	\item  We propose \modelname{}, a novel non-Markovian diffusion model that leverages the full denoising trajectory, leading to more efficient and flexible image generation compared to traditional approaches.
	\item  We propose two key improvements: \modelname{}-AR and \modelname{}-FM, which improve the expressiveness and coherence throughout the non-Markovian generation process.
	\item  \modelname{} achieves competitive performance in both class-conditioned and text-to-image generation tasks, offering a scalable and unified approach for high-quality, controllable image synthesis.
    % \item We propose \modelname{}, a novel model that integrates autoregressive denoising with diffusion processes in a non-Markovian framework, which autoregressively denoise images both spatially and spectrally in a scalable manner. 
    
    % \item \modelname{} achieves competitive performance on standard conditional image generation and text-to-image generation benchmarks.
    
    % \item \modelname{} demonstrates the ability to co-generate text and image data within a unified architecture, simplifying training for text-to-image and class-conditioned generation tasks.
\end{itemize}
%\jg{please check intro}
\section{Background}
\subsection{Diffusion Models}

Diffusion models~\citep{sohl2015deep,ho2020denoising,song2020score} are latent variable models with a fixed posterior distribution, and trained with a denoising objective. These models have gained widespread use in image generation~\citep{rombach2021highresolution, podell2023sdxl,esser2024scaling}. Diffusion models produce the entire image in a non-autoregressive manner through iterative processes. Specifically, given an image $\vx_0 \in\mathbb{R}^{3\times H\times W}$, we define a series of latent variables $\vx_{t}$ ($ t=1,\cdots, T$) with a \emph{Markovian process} which gradually adds noise to the original image $\vx_0$. The transition $q(\vx_t|\vx_{t-1})$ and the marginal $q(\vx_t|\vx_0)$ probabilities are defined as follows, respectively:
\begin{equation}
    q(\vx_{t}|\vx_{t-1}) = \mathcal{N}(\vx_{t}; \sqrt{1-\beta_t}\vx_{t-1},\beta_t\mathbf{I}), \;\; q(\vx_t|\vx_0)=\mathcal{N}(\vx_{t}; \sqrt{\bar{\alpha}_t}\vx_0,\left(1-\bar{\alpha}_t\right)\mathbf{I}),
    \label{eq.DDPM_forward}
\end{equation}
where $\bar{\alpha}_t = \prod_{\tau=1}^t(1-\beta_\tau), 0 < \beta_t < 1$ are determined by the noise schedule.
The model learns to reverse this process with a backward model $p_\theta(\vx_{t-1}|\vx_{t})$, which aims to denoise the image. The training objective for the model is:
\begin{equation}
    \min \mathcal{L}_{\theta}^{\text{DM}} = \mathbb{E}_{t\sim[1,T], \vx_t\sim q(\vx_t|\vx_0)} [\omega_t\cdot \|\vx_\theta(\vx_t, t) - \vx_0\|_2^2],
    \label{eq.DM_loss}
\end{equation}
where $\vx_\theta(\vx_t, t)$ is a time-conditioned denoiser that learns to map the noisy sample $\vx_t$ to its clean version $\vx_0$; $\omega_t$ is a time-dependent loss weighting, which usually uses SNR~\citep{ho2020denoising} or SNR+1~\citep{salimans2022progressive}.
Practically, $\vx_t$ can be re-parameterized with noise- or v-prediction~\citep{salimans2022progressive} for enhanced performance, and can be applied on pixel space~\citep{gu2023matryoshka, saharia2022photorealistic} or latent space, encoded by a VAE encoder~\citep{rombach2021highresolution}.
However, standard diffusion models are computationally inefficient, requiring numerous denoising steps and extensive training data. Moreover, they lack the ability to leverage generation context effectively, hindering scalability to complex scenes and long sequences like videos. %\jg{check}

\subsection{Autoregressive Models}

In the field of natural language processing, Transformer models have achieved notable success in autoregressive modeling~\citep{vaswani2017attention, raffel2020exploring}. Building on this success, similar approaches have been applied to image generation~\citep{parmar2018image, esser2021taming, chen2020generative,yu2022scaling,sun2024autoregressive,team2024chameleon}. Different from diffusion-based methods, these methods typically focus on learning the dependencies among discrete image tokens (e.g., through Vector Quantization~\citep{van2017neural}).
To elaborate, consider an image $\vx\in\mathbb{R}^{3\times H\times W}$. 
The process begins by encoding this image into a sequence of discrete tokens $\vz_{1:N} = \mathcal{E}(\vx)$. These tokens are designed to approximately reconstruct the original image through a learned decoder $\hat{\vx} = \mathcal{D}(\vz_{1:N})$. An autoregressive model is then trained by maximizing the cross-entropy as follows:
\begin{equation}
    \max\mathcal{L}_{\theta}^{\textrm{CE}} = \sum_{n=1}^N \log P_\theta(\vz_n | \vz_{0:n-1}),
    \label{eq.ar}
\end{equation}
where $\vz_0$ is the special start token.
During the inference phase, the autoregressive model is first used to sample tokens from the learned distribution, and then decode them into image space using $\mathcal{D}$.

As discussed in \citet{kilian2024computational}, autoregressive models offer significant efficiency advantages over diffusion models by caching previous steps in memory and enabling the entire generation process to be computed in a single parallel forward pass. This reduces computational overhead and accelerates training and inference. However, the reliance on quantization can lead to information loss, potentially degrading generation quality. Additionally, the linear, step-by-step nature of token prediction may overlook the global structure, making it challenging to capture long-range dependencies and holistic coherence in complex scenes or sequences.
%\jg{add discussion about MAR, VAR, recently...}
\section{\modelname{}}
%In this section, we present \modelname{}, a new class of generative models that is trained with autoregressive denoising objectives. We first derive the formulation of \modelname{} as a special form of non-Markovian diffusion models (\cref{sec.nomad}), then propose methods for improving its performance and scalability (\cref{sec:model}), and finally explore its applications in multi-scale and multi-modal generation (\cref{sec:application}).
\subsection{Non-Markovian Diffusion Formulation}
\label{sec.nomad}
We start by revisiting the basics of diffusion models from the perspective of hierarchical variational auto-encoders~\citep[HVAEs,][]{kingma2013auto,child2020very}. Given a data-point $\vx_0$, a HVAE process $p_\theta$ of a sequence of latent variables $\{\vx_t\}_t$\footnote{We follow the same notation of time indexing $1\ldots T$ in diffusion models for consistency.} by maximizing a evidence lower bound (ELBO):
\begin{equation}
    \max \mathcal{L}^{\text{ELBO}}_{\theta,\phi} = \mathbb{E}_{\vx_{1:T}\sim q_\phi(\vx)} \left[\sum_{t=1}^T\log p_\theta(\vx_{t-1}| \vx_{t:T}) + \log p_\theta(\vx_T) - \log q_\phi(\vx_{1:T}|\vx_0)\right],
\label{eq:elbo}
\end{equation}
where $\vx_0$ is the real data, and $q_\phi$ is a learnable inference model. As pointed out in VDM~\citep{kingma2021variational}, diffusion models can essentially be seen as HVAEs with three specific modifications: 
\begin{enumerate}
    \item A fixed inference process $q$ which gradually adds noise to corrupt data $\vx_0$;
    \item Markovian forward and backward process where $\vx_t$ depends only on $\vx_{t+1}$ (\cref{eq.DDPM_forward});
    \item Noise-dependent loss weighting that reweighs ELBO with a focus on perception.
\end{enumerate}
%To be specific, let's consider a forward diffusion process that defines the diffusion model $q(\vx_t|\vx_{t-1}) = \sqrt{1 - \beta_t^2} \vx_{t-1} + \beta_t \epsilon_t$, where $\beta_t$ is variance schedule and $\epsilon_t \sim \mathcal{N}(0, \mathbf{I})$. It further gives the form $q(\vx_t|\vx_0) = \alpha_t \vx_0 + \sigma_t \epsilon_t$, where $\alpha_t^2 = \prod_{s=1}^t (1-\beta_t^2)$ and $\sigma_t = \sqrt{1-\alpha_t^2}$. 
Only with all above simplifications combined, standard diffusion models can be formulated as \cref{eq.DM_loss} where the generator becomes Markovian $p_\theta(\vx_{t-1}|\vx_{t:T}) = p_\theta(\vx_{t-1}|\vx_t)$ so that one can randomly sample $t$ to learn each transition independently. This greatly simplifies modeling, enabling training models with sufficiently large number of steps (e.g., $T=1000$ for original DDPM~\citep{ho2020denoising}) without suffering from memory issues. 
%First, unlike hierarchical VAE which develops a trainable encoder, encoder in diffusion model does not require training which gradually adds noise to corrupt the data. Second, diffusion model assumes the generation process to be Markovian, namely $x_{t-1}$ only depends on $x_t$ and ignores all the previous steps $x_{t+1:T}$. This leads to a simplified ELBO: $\mathbb{E}_{\vx_{1:T}\sim q(\vx_0)} \left[\sum_{t=1}^T\log p_\theta(\vx_{t-1}| \vx_t) + \log p_\theta(\vx_T) - \log q(\vx_T|\vx)\right]$. Third, diffusion model applies weighted loss across different steps. 
%For example, when using $x_0$ as training objective, the ELBO can be reformulated as $\mathbb{E}_{\vx_{1:T}\sim q(\vx_0)} [\text{SNR}_t \|\hat{\vx}_0(\vx_t;\theta) - \vx_0\|_2^2]$, where $\hat{\vx}_0(\cdot;\theta)$ is the data prediction network that is used to parameterize the denoising process, and the signal-noise ratio (SNR) is defined as $\text{SNR}_t = \alpha_t^2 / \sigma_t^2$. 

In prior research, these aspects are highly coupled, and few works attempt to disentangle them. We speculate that the Markovian assumption might not be a necessary requirement for a high generation quality, as long as the fixed posterior distribution and flexible loss weightings are maintained. As a side evidence, one can achieve reasonable generation with much fewer steps (e.g., $100$) at inference time using non-Markovian HVAE.
On the contrary, the Markovian modeling forces all information compressed solely in the corrupted data from previous noise level which could be an obstacle preventing efficient learning and require more inference steps.

% $q(\vx_t | \vx_0) = \mathcal{N} (\vx_t | \alpha_t \vx_0, \sigma^2 \mathbf{I})$. 
% Its backward process is then given as $q(\vx_{t-1} | \vx_t) = \mathcal{N}(\vx_t | \alpha_t' \vx_{t}, \sigma'^2 \mathbf{I})$
% Formally, a diffusion model defines a forward diffusion process as $q(\vx_t|\vx_{t-1}) = \sqrt{1 - \beta_t^2} \vx_{t-1} + \beta_t \epsilon_t$, where $\beta_t$ is variance schedule and $\epsilon_t \sim \mathcal{N}(0, \mathbf{I})$, which follows the Markovian formulation. Here, we also define signal-noise ratio as $\text{SNR}_t = (1 - \beta_t^2) / \beta_t^2$. 
% $\mathbb{E}_{\vx_{1:T}\sim q(\vx_0)} [\text{SNR}_t \|p_\theta(\vx_t) - \vx_0\|_2^2]$
\paragraph{\underline{NO}n-\underline{MA}rkovian \underline{D}iffusion Models (NOMAD)}
In this paper, we reconsider the original form of generator $p_\theta(\vx_{t-1}|\vx_{t:T})$ of HVAEs while maintaining the modifications, 1 and 3, made by diffusion models. More precisely, we learn the following weighted ELBO loss (\cref{eq:elbo}) with adjustable $\tilde{\omega}_t$:
\begin{equation}
    \max \mathcal{L}^\textrm{NOMAD}_\theta = \mathbb{E}_{\vx_{1:T}\sim q(\vx_0)}\left[
        \sum_{t=1}^T\tilde{\omega}_t\cdot\log p_\theta(\vx_{t-1}|\vx_{t:T})
    \right],
    \label{eq:nomad}
\end{equation}
where  $q$ is the pre-defined inference process. This formulation shares many similarities as autoregressive (AR) models in \cref{eq.ar}, where in our case, each token represents a noisy sample $\vx_t$. Therefore, it is natural to implement such process with autoregressive Transformers~\citep{vaswani2017attention}. 

However, the Markovian inference process of standard diffusion models makes it impossible for the generator $\theta$ to use the entire context except for $\vx_t$. That is to say, even if we initiate our generator as $p_\theta(\vx_{t-1}|\vx_{t:T})$, the model only needs information in $\vx_t$ in order to best denoise $\vx_{t-1}$. 
Therefore, it is critical to design a \textbf{fixed} and \textbf{non-Markovian}\footnote{The term non-Markovian refers to $\vx_t$ is not only related to the previous step $\vx_{t-1}$.} inference process $q(\vx_t|\vx_{0:t-1})$ to sample the noisy sequences $\vx_{1:T}\sim \vx_0$.
The simplest approach is to perform an \textit{independent} noising process:
\begin{equation}
    q(\vx_t|\vx_{0:t-1}) = q(\vx_t|\vx_0) = \mathcal{N}(\vx_t; \sqrt{\gamma_t} \vx_0, (1-\gamma_t)\mathbf{I}), \;\; \forall t \in [1, T],
    \label{eq.n}
\end{equation}
where $\gamma_t$ represents the non-Markovian noise schedule. 
Note that, while \cref{eq.n} may look close to the marginal distribution of the original diffusion models (\cref{eq.DDPM_forward}), the underlying meaning is different as $\vx_t$ is conditionally independent given $\vx_0$. In practice, one can also choose a more complex non-Markovian process as presented in DDIM~\citep{song2021denoising}, and we leave this exploration in future work.
Additionally, we can show the following proposition:
\begin{proposition}
A non-Markovian diffusion process $\{\vx_t\}_t$ with independent noising \(\gamma_t\) has a bijection to a Markovian diffusion process $\{\vy_t\}_t$  with the same number of steps and noise level $\{\bar\alpha_t\}_t$; $\{\vy_t\}_t$ achieves the maximal signal-to-noise ratio when its noise level satisfies \(\frac{\bar{\alpha}_t}{1-\bar{\alpha}_t} = \sum_{\tau=t}^T \frac{\gamma_\tau}{1-\gamma_\tau}\).
\vspace{-10pt}
\label{prop.1}
\end{proposition}
We defer the proof to Appendix~\ref{proof-prop1}. The above proposition indicates that we can carefully choose a set of noise level $\gamma_t$ to model the same information-destroying process as a comparable diffusion baseline, while modeling with the entire generation trajectory.

\begin{figure}[t]
\centering
\setlength{\tabcolsep}{5pt}
    \begin{tabular}{ccc}
    \centering
    \includegraphics[width=0.26\textwidth]{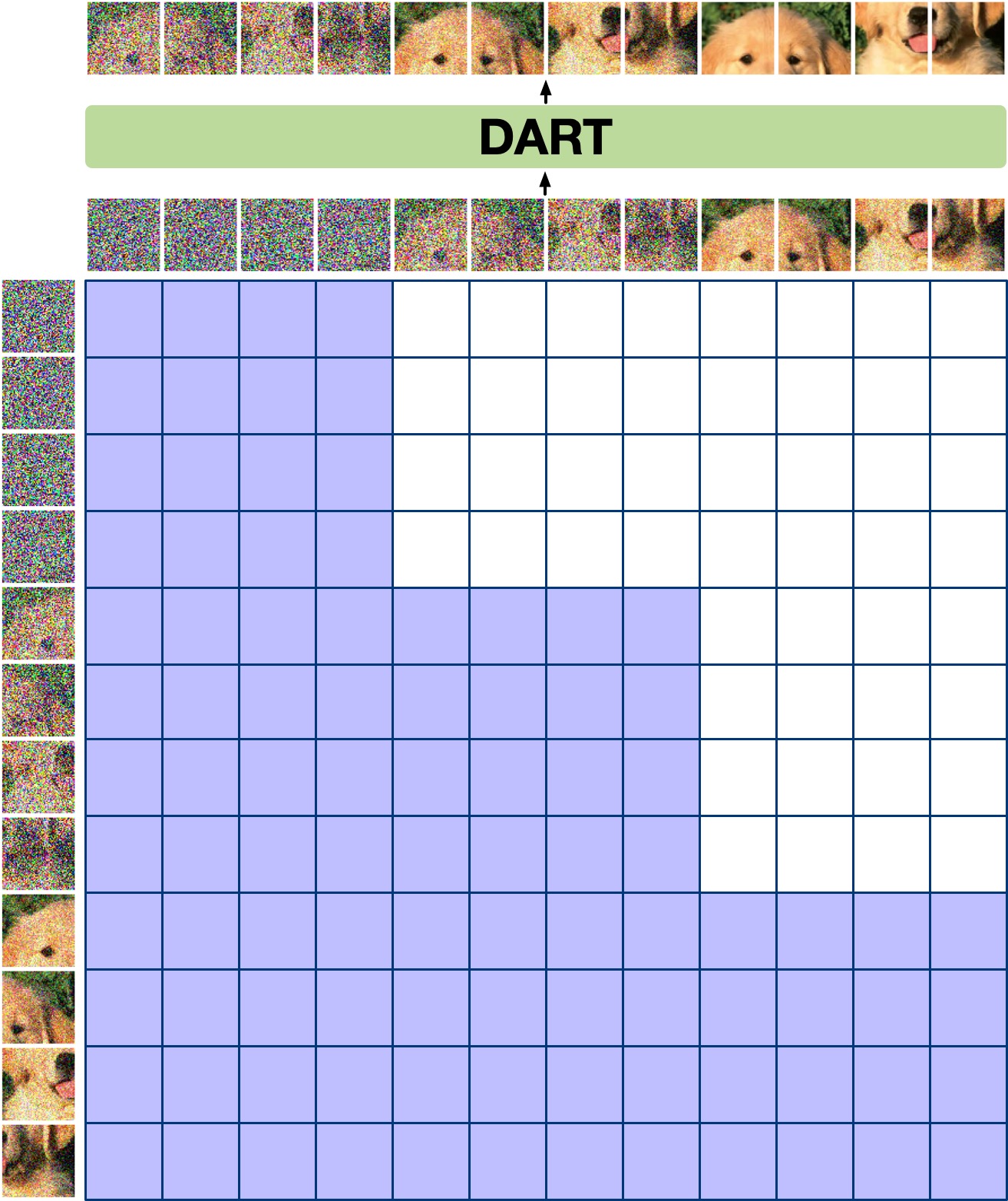} & 
    \includegraphics[width=0.26\textwidth]{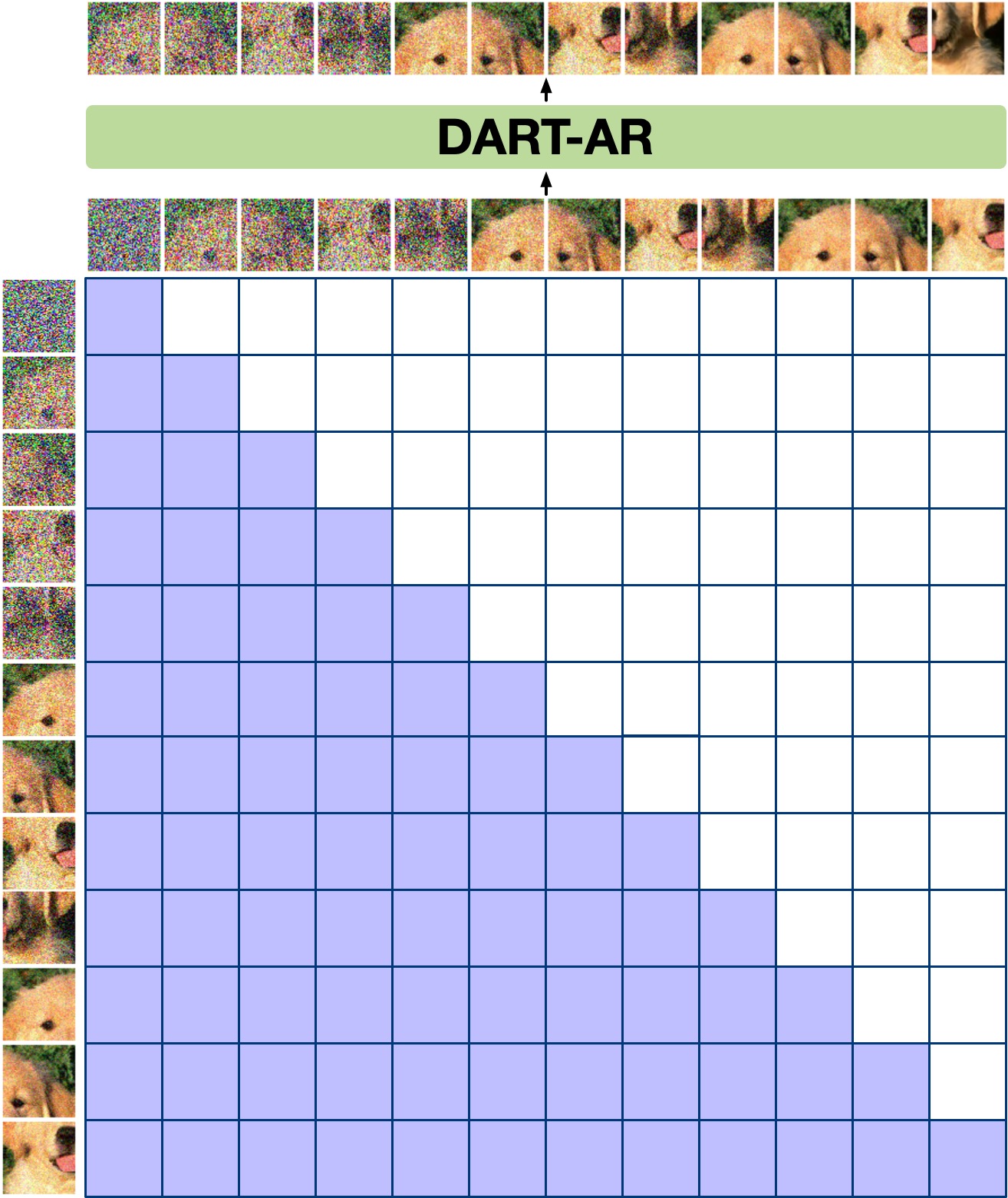} &
    \includegraphics[width=0.33\textwidth]{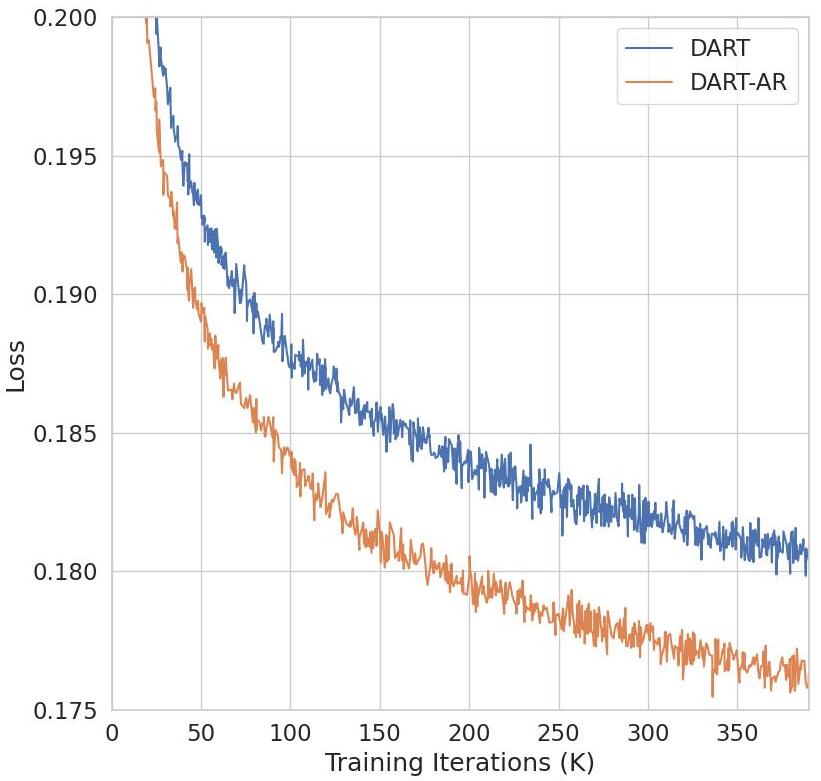}
    \\
    (a) & (b) & (c)
    \end{tabular}
    \caption{Attention masks for (a) \modelname{} and (b) \modelname{}-AR, highlighting their different structures. (c) Comparison of learning curves, demonstrating the superior performance of \modelname{}-AR.}
    \label{fig:attn_mask}
\end{figure}

\subsection{Proposed Methods}
\label{sec:model}

\paragraph{\underline{D}enoising \underline{A}uto\underline{R}egressive \underline{T}ransformer (\modelname{})}
We propose \modelname{} -- a Transformer-based generative model that implements non-Markovian diffusion with an independent noising process (see \cref{fig:dart_pipeline}). 
First, following DiT~\citep{peebles2022scalable}, we represent each image by first extracting the latent map with a pretrained VAE~\citep{rombach2021highresolution}, patchify, and flatten the map into a sequence of continuous tokens $\vx_t\in \mathbb{R}^{K\times C}$, where $K$ is the length, and $C$ is the channel dimension. When considering multiple noise levels,  we concat tokens along the length dimension. 
Then, \modelname{} models the generation as $p_\theta(\vx_{t-1}|\vx_{t:T}) = \mathcal{N}\left(\vx_{t-1}; \sqrt{\gamma_{t-1}}\vx_\theta(\vx_{t:T}), (1-\gamma_{t-1})\mathbf{I}\right)$, where $\vx_\theta(.)$ is a Transformer network that takes in the concatenated sequence $\vx_{t:T} \in \mathbb{R}^{K(T-t)\times C}$, and predicts the ``mean'' of the next noisy image. By combining with \cref{eq.n}, we simplify \cref{eq:nomad} as:
\begin{equation}
    \min \mathcal{L}^{\textrm{\modelname{}}}_\theta = \mathbb{E}_{\vx_{1:T}\sim q(\vx_0)}\left[
        \sum_{t=1}^T\omega_t\cdot\|\vx_\theta(\vx_{t:T}) - \vx_0\|^2_2
    \right],
    \label{eq.dart}
\end{equation}
where we define $\omega_t = \frac{\gamma_{t-1}}{1-\gamma_{t-1}} \tilde{\omega}_t$ to simplify the notation. Similar to standard AR models, training of $T$ denoising steps is in parallel, where computations across different steps are shared. A chunk-based causal mask is used to maintain the autoregressive structure (see \cref{fig:attn_mask}(a)). 

It is evident from \cref{eq.dart} that the objective is similar to the original diffusion objective (\cref{eq.DM_loss}), demonstrating that \modelname{} can be trained as robustly as standard diffusion models. Additionally, by leveraging the diffusion trajectory within an autoregressive framework, \modelname{} allows us to incorporate the proven design principles of large language models~\citep{gpt3, dubey2024llama}.
Furthermore, \cref{prop.1} indicates that we can select $\omega_t$ according to its associated diffusion process. For instance, with SNR weighting~\citep{ho2020denoising}, $\omega_t$ can be defined as $\omega_t = \sum_{\tau=t}^T \frac{\gamma_\tau}{1-\gamma_\tau}$.
%\zdh{refer to prop1?}
    
Sampling from \modelname{} is straightforward: we simply predict the mean $\vx_\theta(\vx_{t:T})$, add Gaussian noise to obtain the next step $\hat{\vx}_{t-1}$, and feed that to the following iteration. Unlike diffusion models, no complex solvers are needed. Similar to diffusion models, classifier-free guidance~\cite[CFG,][]{ho2021classifier} is applied to the prediction of $\vx_\theta$ for improved visual quality.
Additionally, KV-cache is employed to enhance decoding efficiency. %\jg{add cfg}

\paragraph{Limitations of Naive \modelname{}} Unlike diffusion models, which can be trained with a large number of steps $T$, non-Markovian modeling is constrained by memory consumption as $T$ increases. For example, using $T=16$ on $256\times 256$ images will easily create over $4000$ tokens even in the latent space.
This fundamentally limits the modeling capacity on complex tasks such as text-to-image generation.
However, the flexibility of the autoregressive structure allows us to enhance the capacity without compromising scalability. In this work, we propose two methods on top of \modelname{}:

%Our denoising network inherits successful designs of autoregressive language models \citep{gpt3, dubey2024llama} which are based on decoder-only Transformer as shown in \cref{fig:dart_pipeline}. Image are mapped into the feature maps through a pretrained VAE model following DiT~\cite{peebles2022scalable}, and noised are added to the feature maps to corrupt the data. Latent vectors from one corrupted feature map are patchified to convert the spatial input into a sequence of $K$ tokens. For a denoising process of $S$ noise levels, we get a sequence of $K*S$ tokens in total, which are fed into the denoising model sequentially. Unlike diffusion Transformer models like DiT~\cite{peebles2022scalable}, our model implements a decoder-only Transformer, our model allows teacher forcing during training as LLMs and can recursively denoise the corrupted tokens. It also leverages the previous denoising trajectory in generation whereas conventional diffusion model can only see image tokens of single noise level at each time. We also remove adaptive layer normalization (AdaLN) in our design as the timesteps have already been encoded in the positional embeddings of the input tokens. An cross-attention module is also applied to add conditional information (e.g., image class for conditional generation or text embedding for text-to-image generation). 
\begin{figure}[t]
    \centering
    \includegraphics[width=\textwidth]{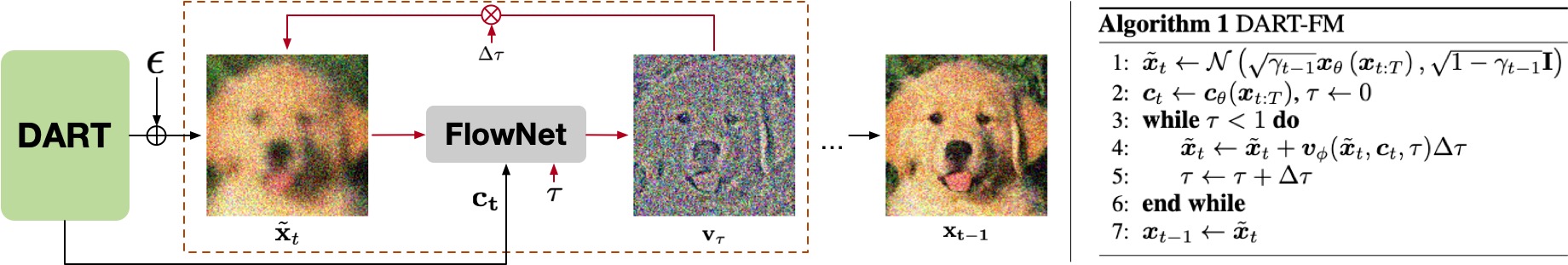}
    \caption{An illustration for the generation process of \modelname{}-FM. }%It can also generate text through next token prediction. Both images and texts are generated in an autoregressive manner. Conditions like text prompt are also added through a cross-attention.}
    \label{fig:dart_fm}
\end{figure}

\paragraph{1. \modelname{} with Token Autoregressive (\modelname{}-AR)}
As similarly discussed by \citet{xiao2021tackling}, the independent Gaussian assumption of $p_\theta(\vx_{t-1}|\vx_{t:T})$ is inaccurate to approximate the complex true distribution of $q(\vx_{t-1}|\vx_{t:T})$, especially when $T$ is small. A straightforward solution to model image  denoising as an additional autoregressive model $p_\theta(\vx_{t-1}|\vx_{t:T}) = \prod_{k=1}^K p_\theta(\vx_{k, t-1}|\vx_{<k, t-1},\vx_{t:T})$:
\begin{equation}
    \min \mathcal{L}^{\textrm{\modelname{}-AR}}_\theta = \mathbb{E}_{\vx_{1:T}\sim q(\vx_0)}\left[
    \sum_{t=1}^T\omega_t
    \sum_{k=1}^K
    \|\vx_\theta(\vx_{<k, t-1}, \vx_{t:T}) - \vx_{k, 0}\|^2_2
    \right],
    \label{eq.dart-ar}
\end{equation}
where $\vx_{1,t-1},\ldots, \vx_{K,t-1}$ are the $K$ flatten tokens of $\vx_{t-1}$. 
The autoregressive decomposition ensures each tokens are not independent, which is strictly stronger than the original \modelname{}. We demonstrate this by visualizing the training curve in \cref{fig:attn_mask} (c).
Training of \modelname{}-AR takes essentially the same amount of computation as standard \modelname{} with two additional modifications at the input and attention masks (see comparison in \cref{fig:attn_mask} (a) (b)).
At sampling time, \modelname{}-AR is relatively much more expensive as it requires $K\times T$ AR steps before it outputs the final prediction.

\paragraph{2. \modelname{} with Flow Matching (\modelname{}-FM)}
The above approach models dependency across tokens, while maintaining Gaussian modeling at each step. Alternatively, we can improve the expressiveness of $p_\theta(\vx_{t-1}|\vx_{t:T})$ by abandoning the Gaussian assumption, similar to \citet{li2024autoregressive}. 
More precisely, we first sample $\tilde{\vx}_t$ as in regular \modelname{}. Then, we recursively apply a continuous flow network, $\vv_\phi(\tilde{\vx}_t, \vc_t, \tau_t)$, over multiple iterations to bridge the gap between $\tilde{\vx}_t$ and $\vx_{t-1}$ (see \cref{fig:dart_fm}). Here, $\vv_\phi(\tilde{\vx}_t, \vc_t, \tau)$ models the velocity field for the probabality flow between the distributions of $\tilde{\vx}_t$ and $\vx_{t-1}$, $\tau \in [0,1]$ denotes the auxiliary flow timestep, and $\vc_t = \vc_\theta(\vx_{t:T})$ represents the features from the last Transformer block, providing contextual information across the noisy image. Consequently, a simple MLP suffices to model $\vv_\phi$, adding only a minimal overhead to the total training cost.
We train $\vv_\phi$ via flow matching~\citep{liu2022flow,lipman2023flow,albergo2023stochastic} due to its simplicity:
\begin{equation}
    \min \mathcal{L}^{\textrm{FM}}_{\phi,\theta} = \mathbb{E}_{\vx_{1:T}\sim q(\vx_0)}\sum_{t=1}^T\mathbb{E}_{\tau \in [0, 1]}\|\vv_\phi((1-\tau)\tilde{\vx}_t + \tau\vx_{t-1}, \vc_t, \tau) - (\vx_{t-1} - \tilde{\vx}_t)\|^2_2,
    \label{eq.fm}
\end{equation}
where $\tilde{\vx}_t = \sqrt{\gamma_{t-1}}\texttt{SG}\left[\vx_\theta(\vx_{t:T})\right] + \sqrt{1-\gamma_{t-1}}\mathbf{\epsilon}$, and $\mathbf{\epsilon} \sim \mathcal{N}(0, \mathbf{I})$. $\texttt{SG}\left[\right]$ is the stop-gradient operator to avoid trivial solutions in optimization.
In practice, we combine \cref{eq.fm} with the original \modelname{} objectives, which can be seen as an additional refinement on top of the Gaussian-based prediction.

\subsection{Multi-resolution Generation}
\label{sec:multi-scale}
\modelname{} (including both the -AR and -FM variants) is a highly flexible framework that can be easily extended and applied in various scenarios with minimal changes in the formulation.
%\paragraph{Progressive Multi-scale Generation}
%Our proposed \modelname{} allows more flexible design for intermediate noisy steps. 
%For example, As one may notice, there are multiple ways to define the denoising order and we only introduce one of them in previous section. For instance, if changing the order to $\{ \vx_T^0, \vx_{T-1}^0, \dots, \vx_{T-K+1}^0, \vx_{T-K}^1, \dots, \vx_0^{K-1}\}$, namely the model first operates the $0$-th token till it is well denoised and then moves to another token. This could be analogous to the setting in recent MAR~\citep{li2024autoregressive}, where clean tokens are generated one after another. 
For example, instead of learning a fixed resolution of images, one can learn a joint distribution of $p_\theta(\{\vx^i_0\}_{i=1}^N)$ where $\vx^i_0\in \mathbb{R}^{K_i\times C}$ is $\vx_0$ with a different resolution. Following the approaches proposed in \citet{gu2022f,gu2023matryoshka,zheng2023learning} for diffusion models, a single \modelname{} — referred to as \emph{Matryoshka-\modelname{}} — can model multiple resolutions by representing each image with its corresponding noisy sequence $\{\vx^k_t\}_t$ separately, then flattening and concatenating these sequences for sequential prediction. 
As a special case shown in \cref{fig:dart_extension} ($\leftarrow$), we  model the NOMAD objective in \cref{eq:nomad} as 
\begin{equation}
    %\log p_\theta (\{\{\vx^i_t\}_{t=0}^{T_i}\}_{i=1}^N) = 
    \max \mathcal{L}^\textrm{Matryoshka}_\theta = \mathbb{E}_{\{\{\vx^i_t\}_{t=0}^{T_i}\}_{i=1}^N\sim q(\vx_0)}\left[
    \sum_{i=1}^N
    \sum_{t=1}^T\tilde{\omega}_t^i\cdot\log p_\theta(\vx_{t-1}^i | \vx^i_{t:T}, \vx^{<i}_{1:T})
    \right],
    \label{eq:multi_scale}
\end{equation}
where the above can be implemented using any \modelname{} variant. Note that $\vx^i_0$ is not directly conditioned on $\vx^{<i}_0$, which not only avoids the need to handle shape changes at the boundaries but also mitigates potential error propagation, a common issue in learning cascaded diffusion~\citep{ho2022cascaded}. All low-resolution information is processed through self-attention.

%the denoising process does not necessarily require a fixed resolution. 
%one can define a sequence of noisy images $\vx_t \in \mathbb{R}^{K_t\times C}$ with variable resolutions, and apply \modelname{} to predict one after another. This can be implemented with any \modelname{} variants
%where tokens of various resolutions will be flatten and concatenated as a long sequence with the original objective (see \cref{fig:dart_extension}). 
%Similar approaches have been explored in diffusion models~\citep{gu2022f,gu2023matryoshka}, however \modelname{} offers a much simpler approach.
By this approach, the model can balance the the number of noise levels with the total number of tokens to achieve better efficiency. Additionally, learning resolutions can be progressively increased by finetuning low-resolution models with extended sequences. % Further implementation details are provided in the Appendix.

% One can define a sequence of corrupted tokens from low-resolution to high-resolution, i.e., $\{ \vx_T^0, \vx_{T-1}^1, \dots, \vx_{T-K_1+1}^{K_1-1}, \vx_{T-K_1}^0, \dots, \vx_0^{K_S-1}\}$. Images of different resolutions contain different number of tokens $\{K_1, K_2, \dots, K_S\}$, the model conducts upsampling besides denoising in autoregressive generation. By this means, the model effectively reduces number tokens and improve learning efficiency. 

%\paragraph{Discrete / Continuous Joint Learning}
\subsection{Multi-modal Generation}
\label{sec:application}
\begin{figure}[t]
    \centering
    \includegraphics[width=\textwidth]{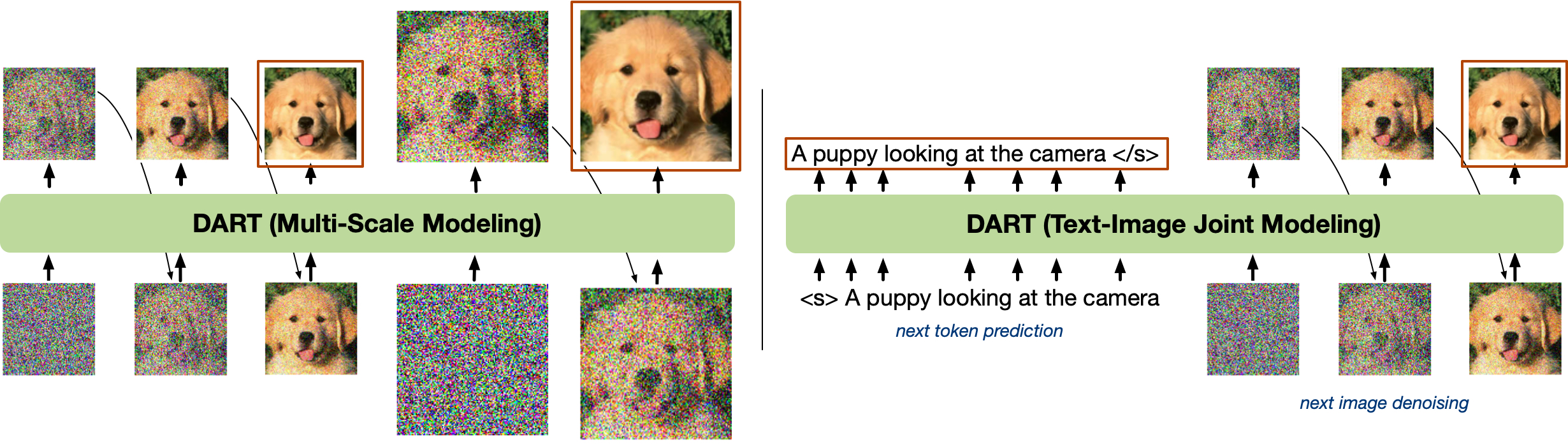}
    \caption{Illustrations of \emph{Matryoshka-\modelname{}} ($\leftarrow$) and \emph{Kaleido-\modelname{}} ($\rightarrow$). By joint training models, \modelname{} can perform various applications such as multi-resolution and multi-modal generation. }%It can also generate text through next token prediction. Both images and texts are generated in an autoregressive manner. Conditions like text prompt are also added through a cross-attention.}
    \vspace{-4pt}
    \label{fig:dart_extension}
\end{figure}

Our proposed framework, built on an autoregressive model, naturally extends to discrete token modeling tasks. This includes discrete latent modeling for image generation~\citep{gu2024kaleido} and multi-modal generation~\citep{team2024chameleon}. By leveraging a shared architecture, we jointly optimize for both continuous denoising and next-token prediction loss using cross-entropy for discrete latents which we call \emph{Kaleido-\modelname{}} considering its architectural similarity as \citet{gu2024kaleido} (see \cref{fig:dart_extension}). To balance inference across modalities, we reweight the discrete loss (\cref{eq.ar}) according to the relative lengths between the discrete and image tokens:
\begin{equation}
    \mathcal{L}^{\textrm{Kaleido}}_\theta = \lambda\mathcal{L}^{\textrm{CE}}_\theta + \mathcal{L}^{\textrm{DART}}_\theta,
\end{equation}
where $\lambda = \frac{\textrm{\# text tokens}}{\textrm{\# image tokens}}$. 
It is important to note that our approach is markedly distinct from several concurrent works aimed at unifying autoregressive and diffusion models within a single parameter space~\citep{zhou2024transfusion,xie2024show,zhao2024monoformer,xiao2024omnigen}. In these efforts, the primary goal is to adapt language model architectures to perform diffusion tasks, without modifying the underlying diffusion process itself to account for the shift in model design. As we have discussed, the Markovian nature of diffusion models inherently limits their ability to leverage generation history, a feature that lies at the core of autoregressive models. In contrast, \modelname{} is designed to merge the advantages of both autoregressive and diffusion frameworks, fully exploiting the autoregressive capabilities. This formulation also allows for seamless integration into LLM pipelines.
%\jg{add some sentence that our method is drastically different from transfusion}

% In the following Experiments section, we show a multimodal version of proposed \modelname{} capable of generating realistic images and text altogether. These indicate that our proposed framework is a generalization of existing denoisng training frameworks which could pave the way for better and generalizable generative modeling. 

%\jg{until here}
\section{Experiments}

% Main model variants: \begin{itemize}
%     \item DART with block size 16
%     \item DART with block size 1
%     \item DART with block size 16 + flow matching 
% \end{itemize}
\begin{figure}[t]
    \centering
    \includegraphics[width=0.96\linewidth]{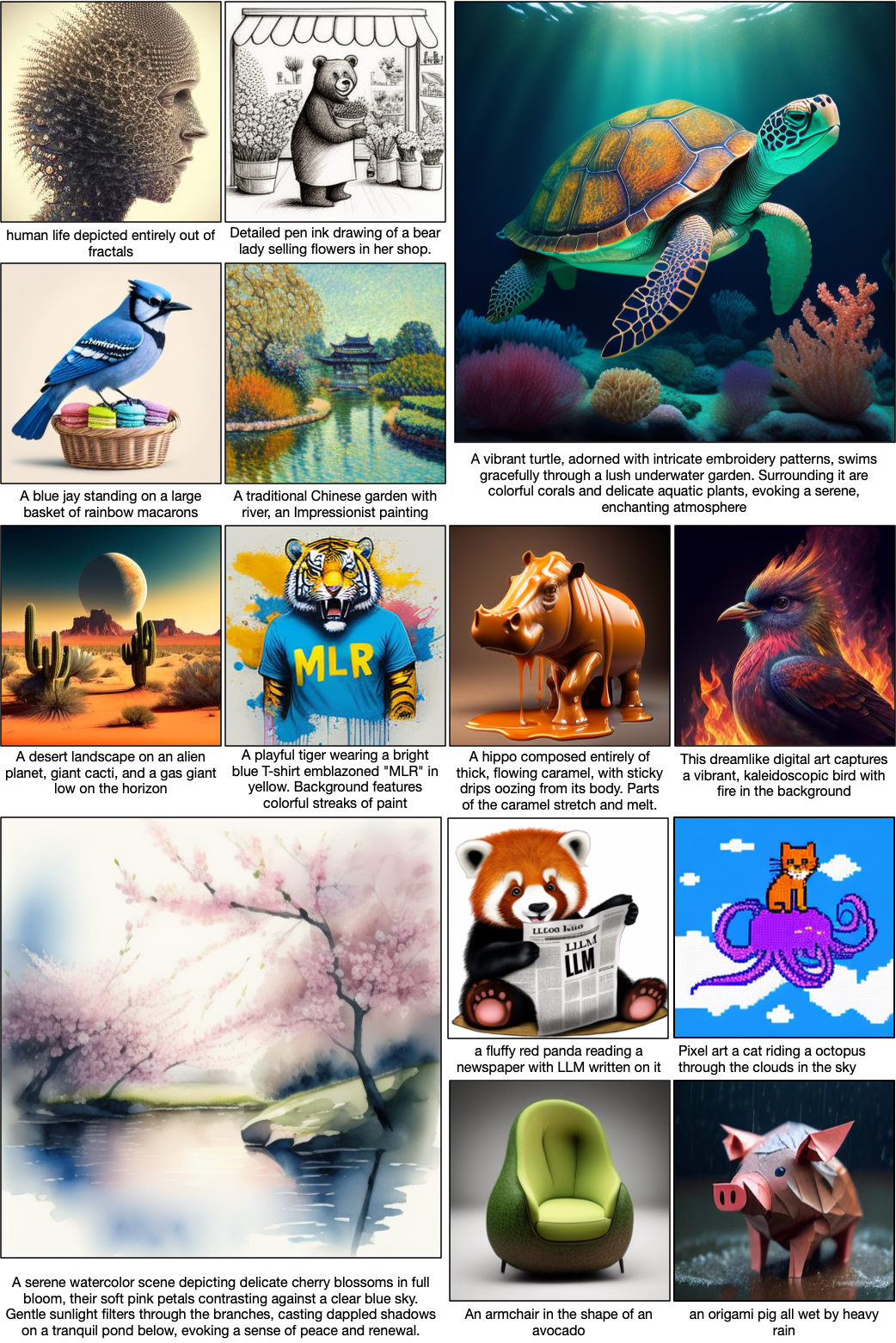}
    \caption{\small Samples generated by text-to-image \modelname{}-FM at $256\times 256$ and $512\times 152$ pixels}
    \vspace{-15pt}
    \label{fig:main_results}
\end{figure}
\begin{figure}[t]
    \centering
    \includegraphics[width=\linewidth]{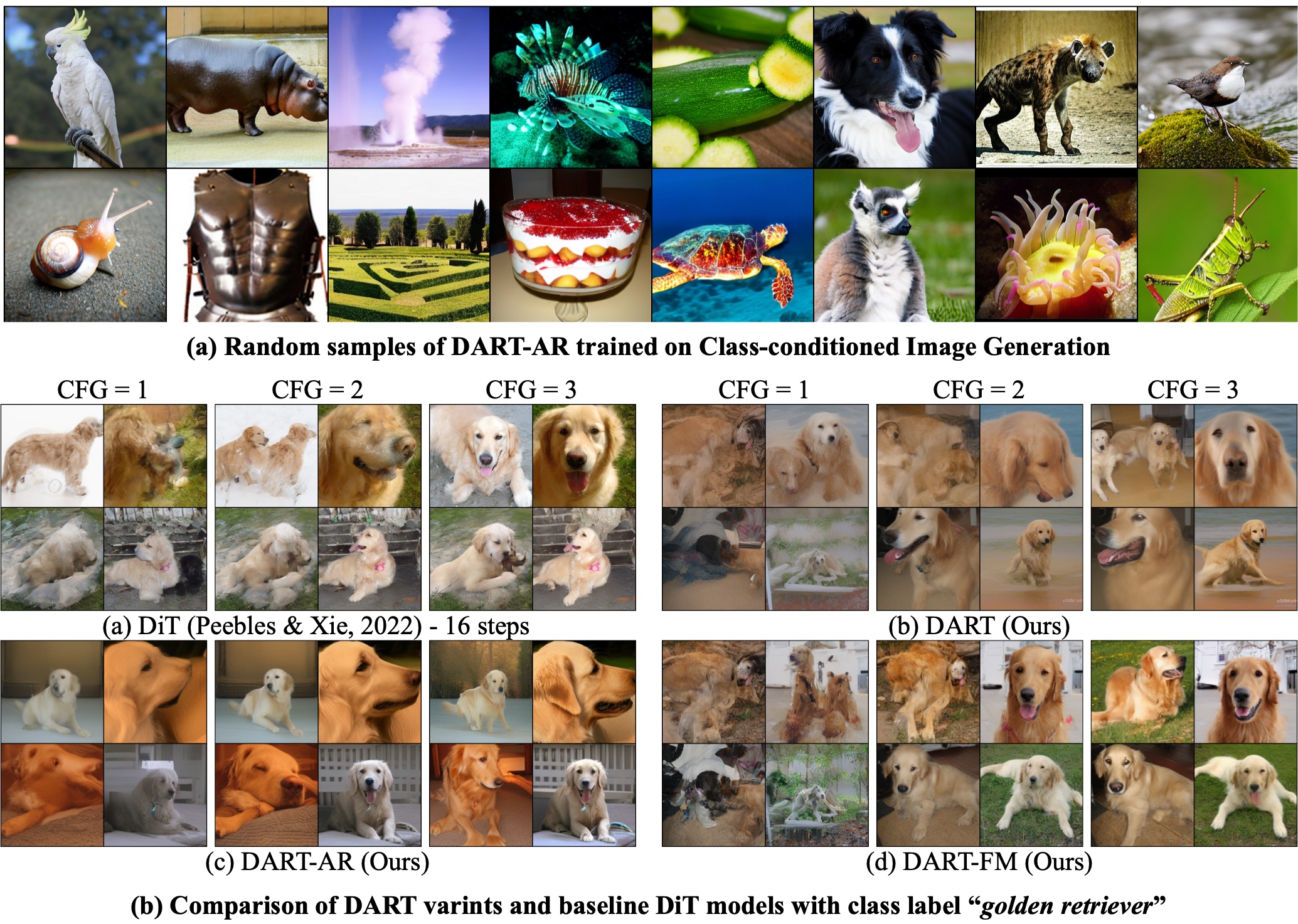}
    \vspace{-15pt}
    \caption{\small Comparison for class-conditioned image generation with \modelname{} and baselines trained on ImageNet.}
    \label{fig:imagenet_compare}
\end{figure}
\subsection{Experimental Settings}

\paragraph{Dataset} We experiment with \modelname{} on both class-conditioned image generation on ImageNet~\citep{Deng2009ImageNet:Database} and text-to-image generation on CC12M~\citep{changpinyo2021cc12m}, where each image is accompanied by a descriptive caption. All images are center-cropped and resized.
All models, except those used for Matryoshka-\modelname{} finetuning, are trained to synthesize images at $256 \times 256$ resolution, while the latter are trained at $512 \times 512$. For multimodal generation tasks, we  augment CC12M with synthetic captions as the discrete ground-truth.
%In addition,  we also adopt the validation set of COCO2017 for testing zero-shot text-to-image generation.

\vspace{-5pt}\paragraph{Evaluation}  In line with prior works, we report Fr\'echet Inception Distance (FID)~\citep{fid} %and Inception Score (IS) 
to quantify the the realism and diversity of generated images. %Besides, we employ Recall~\citep{kynkaanniemi2019improved} to measure the diversity of samples. 
For text-to-image generation, we also use the CLIP score~\citep{hessel2021clipscore} to measure how well the generated images align with the given text instructions. To assess the zero-shot capabilities of the models, we report scores based on the MSCOCO 2017~\citep{Lin2014} validation set.

\vspace{-5pt}\paragraph{Architecture}
Following \cref{sec:model}, we experimented with three variants of the proposed model: the default \modelname{}, along with two enhanced versions, \modelname{}-AR and \modelname{}-FM. All variants are implemented using the same Transformer blocks for consistency. As illustrated in \cref{fig:dart_pipeline}, our design is similar to \citet{dubey2024llama}, incorporating rotary positional encodings~\citep[RoPE,][]{su2024roformer} within the self-attention layers and SwiGLU activation~\citep{shazeer2020glu} in the FFN layers. For class-conditioned generation, we follow~\citep{peebles2022scalable}, adding an AdaLN block to each Transformer block to integrate class-label information. For text-to-image generation, we replace AdaLN with additional cross-attention layers over pretrained T5-XL encoder~\citep{raffel2020exploring}. Since \modelname{} uses a fixed noise schedule, there is no need for extra time embeddings as long as RoPE is active. 
In addition, for \modelname{}-FM, we incorporate a small flow network, implemented as a 3-layer MLP, which increases the total parameter count by only about 1\%. 
% Further details on the hyper-parameters used for this configuration are provided in the Appendix.

\vspace{-5pt}\paragraph{Training \& Inference}
\cref{prop.1} not only makes a connection between \modelname{} and diffusion models, but also allows us to define the noise schedule based on any existing diffusion schedule $\{\bar{\alpha}_t\}_t$, which can be inversely mapped to the \modelname{} schedule $\{\gamma_t\}_t$ using the bijection. In this paper, we adopt the cosine schedule $\bar{\alpha}_t=\cos{\left({\pi }/{2}\cdot{t}/{T}\right)}$. We set $T=16$ while $K=256$\footnote{$K=256$ from encoding $256 \times 256$ images with StableDiffusion v1.4 VAE (\url{https://huggingface.co/stabilityai/sd-vae-ft-ema}) with a patch size of 2 and $C=16$ channels.} throughout all experiments unless otherwise specified.
We train all models with a batch size of 128 images, resulting in a total of 0.5M image tokens per update. 
For finetuning Matryoshka-\modelname{}, we additionally set the high-resolution part $T=4$ with $K=1024$.
We use the AdamW optimizer~\citep{loshchilov2017decoupled} with a cosine learning rate schedule, setting the maximum learning rate to $3\textrm{e-}4$. 
During inference, we enable KV-cache with memory pre-allocated based on the total sequence length. \modelname{} performs the same number of steps for inference as in training. For the FM variants, we additionally apply $100$ flow matching steps between each autoregressive step, following \citet{li2024autoregressive}. More implementation details can be found in Appendix~\ref{app:implement}.

\subsection{Results}
\begin{figure}[t]
\centering
\setlength{\tabcolsep}{0pt}
    \begin{tabular}{ccc}
    \centering
    \includegraphics[width=0.333\textwidth]{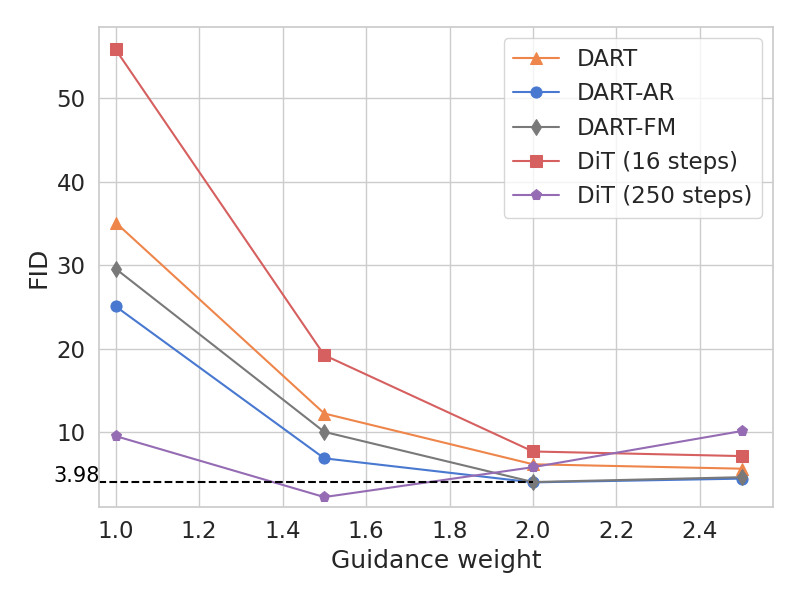} & 
    \includegraphics[width=0.333\textwidth]{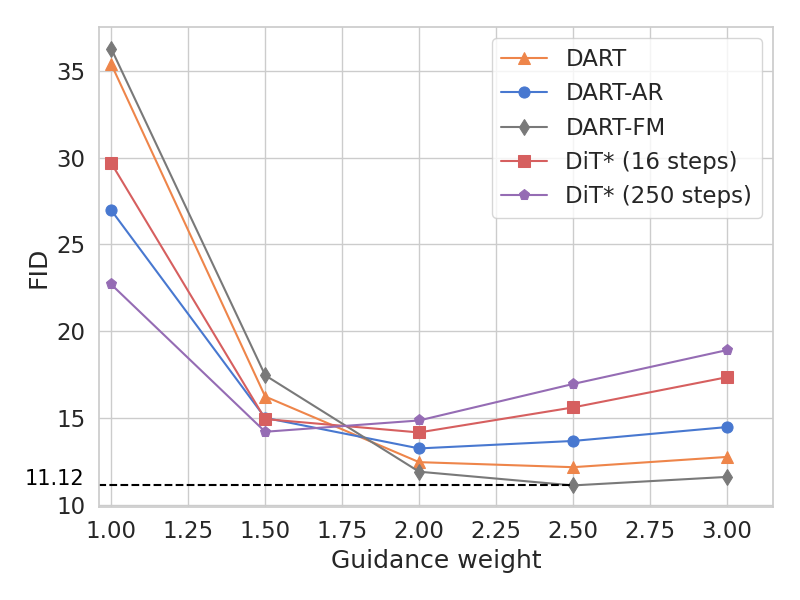} &
    \includegraphics[width=0.333\textwidth]{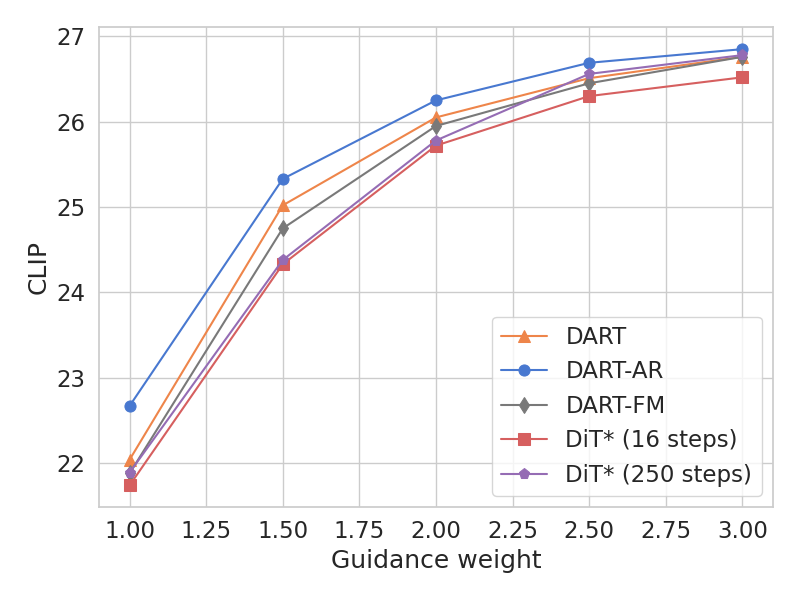}
    \\
    {\small (a) FID50K on ImageNet} &  {\small(b) FID30K on COCO}&  {\small(c) CLIP score on COCO}
    \end{tabular}
    \caption{\small Comparison of \modelname{}, \modelname{}-AR, \modelname{}-FM and baseline models with different CFG guidance scale on different benchmarks. * denotes models implemented and trained by us.}
    \label{fig:fid_clip}
\end{figure}

\paragraph{Class-conditioned Generation}
We report the FID scores for conditional ImageNet generation in \cref{fig:fid_clip}(a), following the approach of previous works. In line with \cite{li2024autoregressive}, we apply a linear CFG scheduler to \modelname{} and its variants. As shown in the figure, both -AR and -FM variants consistently outperform the default \modelname{} across all guidance scales, demonstrating the effectiveness of the proposed improvement strategies. We also compare our methods with DiT~\citep{peebles2022scalable}, using both 16 sampling steps (to match \modelname{}) and 250 steps (the suggested best setting). Notably, \modelname{}-AR achieves the best FID score of \textbf{3.98} among all variants and significantly surpasses DiT when using 16 steps, highlighting its advantage in leveraging generative trajectories, particularly when the number of sampling timesteps is limited. Not surprisingly, DiT with 250 steps performs better than \modelname{}. However, it is important to note that the official DiT model is trained for 7M steps, which is substantially more training iterations than those used for \modelname{}.
%DiT is also sensitive to the guidance scale in CFG, for example, with scale as $2.0$, DiT performs almost the same as \modelname{} and is worse than \modelname{}-AR and \modelname{}-FM. 

\Cref{tb:imagenet} lists the performance of our proposed \modelname{} in comparison to recent generative models on ImageNet-256. The reproduced results use the official codebase and checkpoint and we follow the best performing cfg scale reported in the original papers. We report reproduced results of DiT~\citep{peebles2022scalable}, SiT~\citep{ma2024sit}, and MAR~\cite{li2024autoregressive} with 16 sampling times which is the same as our vanilla DART. 
Our model achieves competitive performance when compared with diffusion models like LDM~\citep{rombach2021highresolution} and AR models like VQGAN~\citep{esser2021taming} and RQ-Transformer~\citep{lee2022autoregressive}. Admittedly, there is a gap between DART and SOTA visual generative models (like VAR~\citep{tian2024visual} and MAR~\citep{li2024autoregressive}). However, we want to point out that many baselines are trained with significantly more FLOPs. For example, DiT~\citep{peebles2022scalable} is trained for 7M iterations whereas DART is only trained for 500k iterations. Also, baselines like VAR and MAR employ larger models than our DART. In particular, VAR deploys a 2B model while the largest DART model is approximately 800M. Besides, in our reproduced results, when using only 16 sampling steps as the setting of our vanilla DART, our model show significantly better performance than DiT, SiT and MAR. 
Also, we report MAR-AR~\citep{li2024autoregressive}, a variant of MAR which generate tokens in an autoregressive manner instead of masked modeling which is applied in standard MAR models. DART which generates samples through autoregressive denoising shows better performance than MAR-AR. 
These results further validate the effectiveness of leveraging the whole denoising trajectory. 

\begin{table}[t]
\caption{Generative models on class-conditional ImageNet $256 \times 266$. *: reproduced from official codebase and checkpoints. }
\label{sample-table}
\begin{center}
\resizebox{0.8\textwidth}{!}{
\begin{tabular}{c l c c c c}
\toprule
Type & Model & FID$\downarrow$ & IS$\uparrow$ & \#params & Steps \\
\midrule
\multirow{4}{*}{Diff.} & ADM~\citep{dhariwal2021diffusion} & 10.94 & 101.0 & 554M & 250 \\
& CDM~\citep{ho2022cascaded} & 4.88 & 158.7 & - & 8100\\
& LDM~\citep{rombach2021highresolution} & 3.60 & 247.7 & 400M & 250 \\
& DiT~\citep{peebles2022scalable} & 2.27 & 278.2 & 675M & 250 \\
& SiT~\citep{ma2024sit} &  2.06 & 277.5 & 675M & 250 \\
\midrule
\multirow{4}{*}{AR} & VQGAN~\citep{taming} & 15.78 & 74.3 & 1.4B & 256 \\
& RQTran~\citep{lee2022autoregressive} & 3.80 & 323.7 & 3.8B & 68 \\
& MAR-AR~\citep{li2024autoregressive} & 4.69 & 244.6 & 479M & 256 \\
& MAR~\citep{li2024autoregressive} & 1.55 & 303.7 & 943M & 256 \\
& VAR~\citep{tian2024visual} & 1.73 & 350.2 & 2.0B & 10 \\
\midrule
\multirow{3}{*}{Reprod.} & DiT*~\citep{peebles2022scalable} & 19.52 & 125.9 & 675M & 16 \\
% & DiT*~\citep{peebles2022scalable} (cfg=2.0) & 7.75 & 225.16 & 675M & 16 \\
& SiT*~\citep{ma2024sit} & 6.98 & 122.9 & 675M & 16 \\
% & SiT*~\citep{ma2024sit} (cfg=2.0) & 3.80 & 341.72 & 675M & 16 \\
& MAR*~\citep{li2024autoregressive} & 6.37 & 221.3 & 943M & 16 \\
\midrule
\multirow{3}{*}{Ours}& DART & 5.62 & 231.7 & 812M & 16 \\
& DART-AR & 3.98 & 256.8 & 812M & 4096 \\
& DART-FM & 3.82 & 263.8 & 820M & 16 \\
\bottomrule
\end{tabular}
}
\end{center}
\label{tb:imagenet}
\end{table}

\cref{fig:imagenet_compare} also presents examples of generated results on ImageNet from all models using 16 sampling steps. \modelname{}-FM tends to produce sharper images with higher fidelity at higher CFG values. In contrast, \modelname{}-AR demonstrates an ability to generate more realistic samples at lower CFG values when compared to both the baselines and other variants. More samples are shown in Appendix~\ref{app:additional_samples}.

\begin{figure}[t]
\centering
\setlength{\tabcolsep}{0pt}
    \begin{tabular}{ccc}
    \centering
    \begin{minipage}[t]{0.333\textwidth} % Align table at the top
    \raisebox{50pt}{%
        \small
        \setlength{\tabcolsep}{4pt} % Adjust this value as needed
        \begin{tabular}{lrr}
        \toprule
        Model & Gflops & Speed (s) \\
        \midrule
        \modelname{} & 2.157  & 0.075\\
        \modelname{}-AR & 2.112 & 7.44 \\
        \modelname{}-FM & 2.177 & 0.32 \\
        \midrule
        DiT (16) & 2.073 & 0.065\\
        DiT (256) & 32.384 & 0.84\\
        \bottomrule
        \end{tabular} 
        \setlength{\tabcolsep}{0pt}
        }
        \end{minipage}
        & 
    \includegraphics[width=0.333\textwidth]{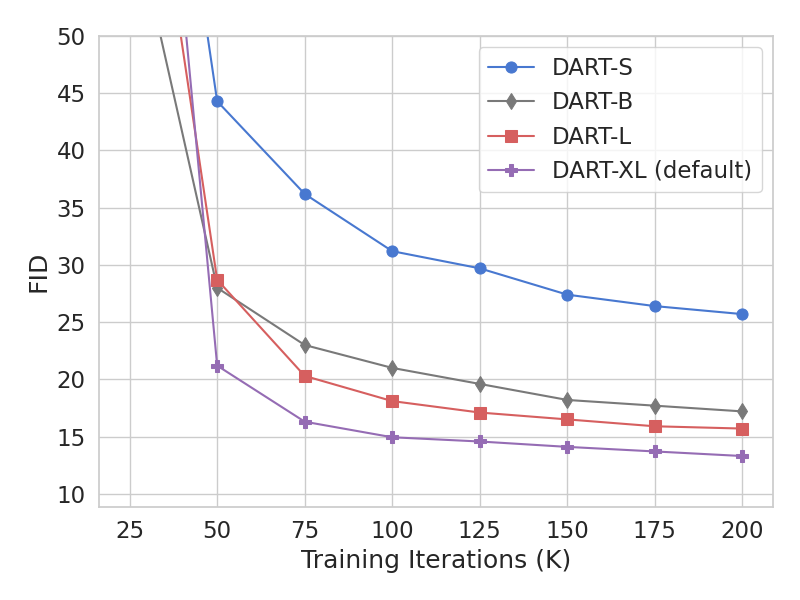} &
    \includegraphics[width=0.333\textwidth]{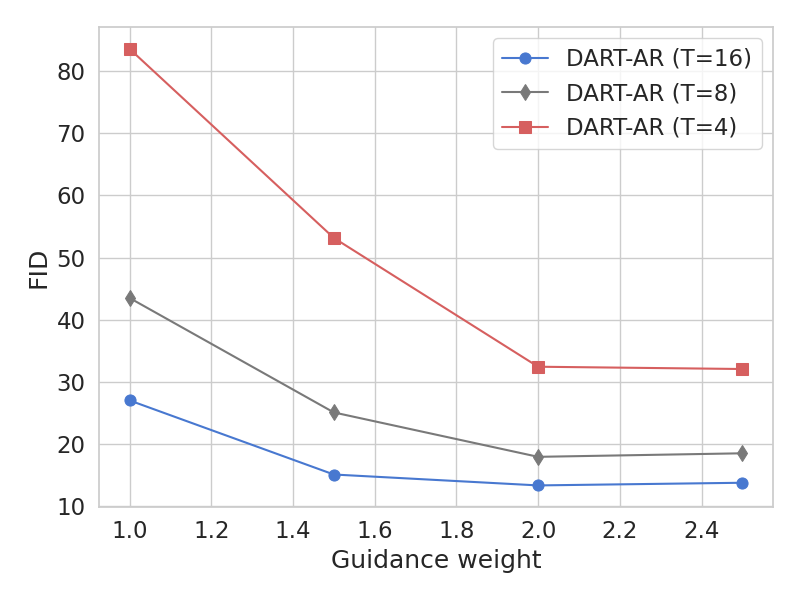}
    \\
    {\small (a) Inference speed \& flops} &  {\small(b) Comparison of model sizes}&  {\small(c) Comparison of \# of steps}
    \end{tabular}
    \caption{\small (a) Inference flops and speed (caulated as second per image) of different models. (b) Performance of \modelname{} of different sizes. (c) Effect of number of noise levels on \modelname{}.} %\textcolor{blue}{TODO: specify benchmark setting \& change time to sample/sec}}
    \label{fig:fid_sizes_steps}
    \vspace{-10pt}
\end{figure}
\vspace{-5pt}
\paragraph{Text-to-Image Generation}
To demonstrate the capability of \modelname{} at scale, we train the model for text-to-image generation. 
%We here report FID and CLIP score~\citep{hessel2021clipscore} (see \cref{fig:fid_clip}(b)(c)) where the latter is a commonly used metric to measure how well text-to-image generative models follow text instructions. 
We also implement an in-house DiT with cross attention to text condition in comparison to our models, where we evaluate the performance on both $16$ and $250$ steps. 
%Here, we either set training and inference timesteps for DiT as 16 (DiT* 16 steps) or set them to be 1000 and 250 respectively (DiT* 250 steps). 
As shown in \cref{fig:fid_clip}, while both -AR and -FM variants still show clear improvements against the default \modelname{}, FM achieves the best FID of \textbf{11.12}, indicating its ability of handling diverse generation tasks.

To validate the scalability of high-resolution image generation, we further finetune \modelname{} by jointly modeling $256\times 256$ and $512\times 512$ resolutions within a single autoregressive process (\cref{sec:multi-scale}). Sampled results are shown in \cref{fig:main_results} and Appendix~\ref{app:additional_samples}.

\vspace{-5pt}
\paragraph{Efficiency}
We compare both the actual inference speed (measured by wall-clock time with batch size 32 on a single H100) as well as the theoretical computation (measured by GFlops) in \ref{fig:fid_sizes_steps}(a). Since \modelname{}, \modelname{}-AR, \modelname{}-FM share the same encoder-decoder Transformer architecture, their flops are roughly the same. However, \modelname{}-AR has high wall clock inference time due to its large autoregressive steps, which have not been well parallelized in our current implementation. Integrated with recent advances in autoregressive LLMs, \modelname{}-AR can be deployed in a more efficient and we leave this for future investigations. \modelname{}-FM also have a inference time overhead due to the iterations of flow net in inference. Compared to \modelname{}, DiT has less flops for a single pass. However, it requires sufficient many number of iterations to generate high quality samples. \modelname{} has comparable flops and inference time as DiT with 16 sampling timesteps, while \modelname{} achieves better performance than DiT (16) on ImageNet and COCO, showing the efficiency benefits of \modelname{}.

\begin{figure}[t]
    \centering
    \includegraphics[width=\linewidth]{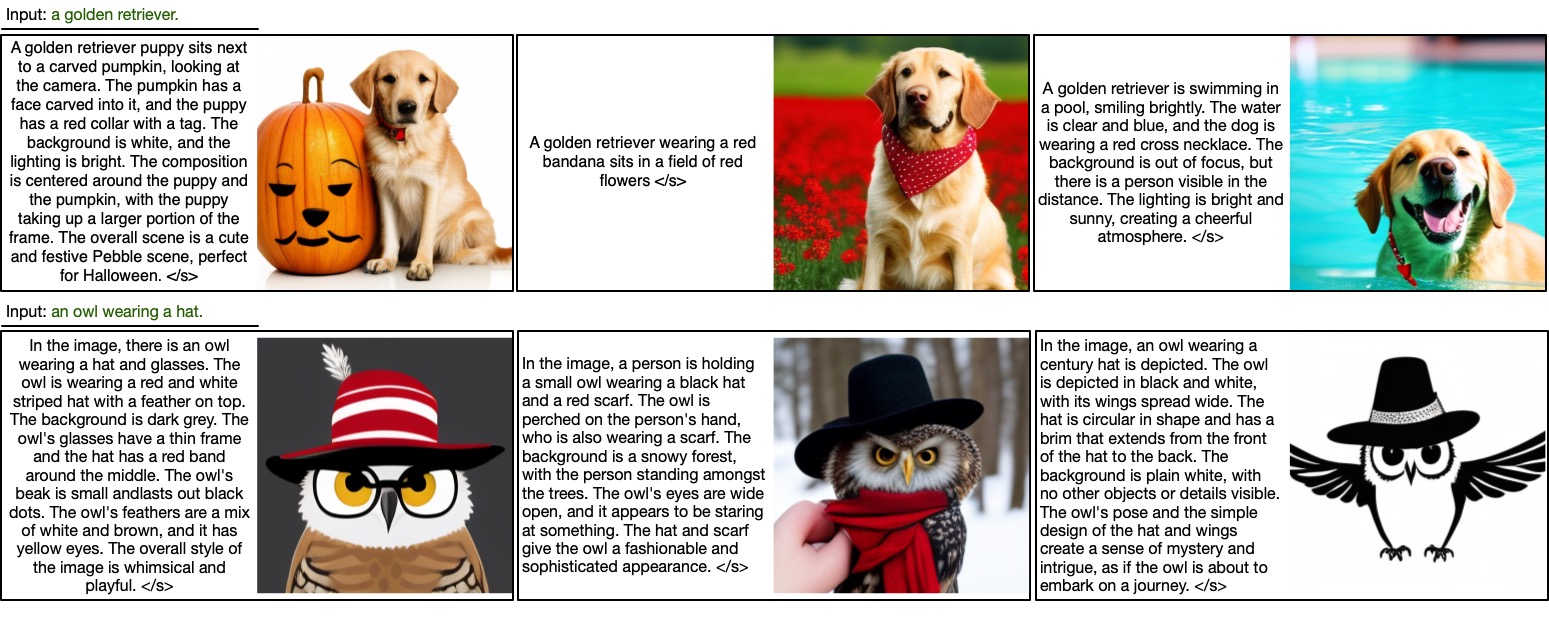}
    \caption{Examples of multi-modal generation with Kaleido-\modelname{}. }\vspace{-10pt}
    \label{fig:kaleido}
\end{figure}
%\jg{@yuyang}
%\subsection{Ablation Study}

% \begin{itemize}
%     \item different step sizes e.g. 4, 8, 16
%     \item different block size (optional)
%     \item different patch order
%     \item Multi-resolution
% \end{itemize}
%\paragraph{Classifer-free guidance (CFG)} \modelname{} can adapt CFG seamlessly by replacing the text conditional with null token in inference. We first investigate the optimal setting of applying CFG in inference. We employ three CFG settings including constant, linear, and CFG++. Constant CFG applies a constant guidance scale in generating all the image tokens. Following \cite{li2024autoregressive}, linear CFG applies a linearly increasing guidance scale sequentially. CFG++~\citep{chung2024cfg++} is a recent improved guidance technique which defines a bell-shape guidance scale that achieves maximal value in the middle timestep. Fig.~\ref{} shows the effect of different CFG strategies. As shown, linear CFG and CFG++ greatly improves the CLIP score compared with the constant counterpart. Therefore, we employ linear CFG in this work if not mentioned otherwise. 

\vspace{-5pt}
\paragraph{Scalibility} We show the scalability of our \modelname{} by training models of different sizes including small (S), base (B), large (L), and extra large (XL) on CC12M, where the configurations are listed in Appendix \ref{app:config}. Figure~\ref{fig:fid_sizes_steps}(b) illustrates how the performance changes as model size increase. Across all the four models, CLIP score significantly improves by increasing the number of parameters in \modelname{}. Besides, the perform steadily increases as more training iterations are applied. This demonstrates that our proposed generative paradigm benefits from scaling as previous generative models like diffusion~\citep{peebles2022scalable} and autoregressive models~\citep{li2024autoregressive}.

\vspace{-5pt}
\paragraph{Effects of noise levels.} We experiment with different noise levels, where we vary the number of total noise levels in our denoising framework (Figure~\ref{fig:fid_sizes_steps}(c)). A total of $T=4$ or $8$ noise levels are trained in comparison with standard $16$ noise levels. Not surprisingly, less noise levels lead to gradually degraded performance. However, with as few as 4 noise levels, \modelname{} can still generate plausible samples, indicating the capabilities of deploying \modelname{} in a more efficient way. Variants in noise levels provides a potential option for finding the optimal computation and performance trade-off especially when compute is limited. 

\vspace{-5pt}
\paragraph{Multimodal Generation} We showcase the capabilities of our proposed method in the joint generation of discrete text and continuous images, as introduced in \cref{sec:application}. \cref{fig:kaleido} provides examples of multimodal generation using the Kaleido-\modelname{} framework. Given an input, the model generates rich descriptive texts along with corresponding realistic images, demonstrating its ability to produce diverse samples with intricate details. Notably, unlike \citet{gu2024kaleido}, our approach processes both text and images through the same model, utilizing a unified mechanism to handle both modalities. This unified framework can potentially be integrated into any multimodal language models.
%For instance, given ``golden retriever'' as input, the model generates pictures of different backgrounds and decorations. 

% \subsection{Text image co-generation}
%\input{sections/related_work}
\section{Conclusion}
We presented \modelname{}, a novel model that integrates autoregressive denoising with non-Markovian diffusion to improve the efficiency and scalability of image generation. By leveraging the full generation trajectory and incorporating token-level autoregression and flow matching, \modelname{} achieves competitive performance on class-conditioned and text-to-image tasks. This approach offers a unified and flexible solution for high-quality visual synthesis. 

Our current model is restricted by the number of tokens in the denoising process. One direction is to explore more efficient architectures for long context modeling~\citep{gu2023mamba,yan2024diffusion} which enables application to problems like video generations. Also, current work only conduct preliminary investigates on multimodal generation. Future work may train a multimodal generative model based on the framework of the decoder-only language model to handle a wide variety of problems with one unified model. 

\bibliography{references}
\bibliographystyle{iclr2025_conference}
\newpage
\appendix

\section{Proof of \cref{prop.1}}
%Regarding Proposition~\ref{prop.1}:
\begin{proof}
\label{proof-prop1}
Due to $q(\vx_t|\vx_0) = \mathcal{N}(\vx_t; \sqrt{\gamma_t} \vx_0, (1-\gamma_t)\mathbf{I})$, we can write $\vx_t =\sqrt{\gamma_t} \vx_0 + \sqrt{1-\gamma_t}\veps_t$.

With certain coefficient $\{\lambda^t_s\}_{s=t}^T$, let us define: 
\begin{align}
    \vy_t &\triangleq 
    %\begin{cases}
     \sum_{s=t}^T \lambda^t_s \vx_s,  %& \text{if } t < T \\
     %\vx_t, & \text{if } t = T
    %\end{cases}
    %\;\; \emph{s.t.}  \;\; \sum_{s=t}^T\lambda^t_s = 1
    \label{eq.y}
\end{align}

We study when the signal-to-noise ratio of $\vy_t$ achieves its maximal value:
\begin{align}
    \vy_t &= \left(\sum_{s=t}^T \lambda^t_s \sqrt{\gamma_s}\right)\vx_0 + \sum_{s=t}^T \lambda^t_s \sqrt{1-\gamma_s}\veps_s \label{eq.11} \\
    \Rightarrow \text{SNR}(\vy_t) &= \frac{\left(\sum_{s=t}^T \lambda^t_s \sqrt{\gamma_s}\right)^2}{\sum_{s=t}^T {\lambda^t_s}^2 (1-\gamma_s)} \le \sum_{s=t}^T\frac{\gamma_s}{1-\gamma_s},
\end{align}
which follows from Titu's lemma and the Cauchy-Schwarz inequality. It becomes an equality when:
\begin{align}
\lambda^t_s\propto \frac{\sqrt{\gamma_s}}{1-\gamma_s}, \forall s\in [t, T]
\label{eq.13}
\end{align}

% Given the normalization constraint in \cref{eq.y}, a deterministic solution can be obtained:
% \begin{align}
% \lambda^t_s = \frac{\sqrt{\gamma_s}}{1-\gamma_s} \bigg/ \sum_{s=t}^T\left[\frac{\sqrt{\gamma_s}}{1-\gamma_s}\right]
% \end{align}

Next, we demonstrate that $\{\vy_t\}_t$ follows a Markov property when achieving the maximal signal-to-noise ratio for each $\vy_t$. 
From \cref{eq.13}, let 
$\lambda_s^t = \rho_t \frac{\sqrt{\gamma_s}}{1-\gamma_s}, \rho_t > 0$, and $\eta_t = \frac{\gamma_t}{ 1-\gamma_t}$, $\bar{\eta}_t = \sum_{s=t}^T\eta_s$, we have:
\begin{align}
    \vy_t &= \rho_t \left(\bar{\eta}_t\vx_0 + \sum_{s=t}^T\sqrt{\eta_s}\veps_s\right) \\
    &= \rho_t \left(\bar{\eta}_t\vx_0 + \sqrt{\bar{\eta}_t}\veps'_t\right),
\end{align}
where we use $\veps'_t \sim \mathcal{N}(0, \mathbf{I}) $ to equivalently simplify the noise term. When $\rho_t = 1\big/\sqrt{\bar{\eta}^2_t + \bar{\eta}_t}$, $\{\vy_t\}_t$ is variance preserving (VP). Next, let us assume 
\begin{align}
    \hat{\vy} &= 
    \frac{\rho_{t+1}\bar{\eta}_{t+1}}{\rho_{t}\bar{\eta}_{t}}\vy_t + \sigma \veps, \veps \sim \mathcal{N}(0, \mathbf{I}) \\
    &= \rho_{t+1}\bar{\eta}_{t+1}\vx_0 + \rho_{t+1}\bar{\eta}_{t+1}/\sqrt{\bar{\eta}}_t\veps'_t + \sigma\veps \\
    &= \rho_{t+1}\left(\bar{\eta}_{t+1}\vx_0 + \sqrt{\frac{\bar{\eta}_{t+1}^2}{\bar{\eta}_t} + \frac{\sigma^2}{\rho^2_{t+1}}}\veps''
    \right),
\end{align}
where we use $\veps''$ to replace the noise term.
So if we let $\hat{\vy}$ match the distribution of $\vy_{t+1}$, then 
\begin{align}
    &\frac{\bar{\eta}_{t+1}^2}{\bar{\eta}_t} + \frac{\sigma^2}{\rho^2_{t+1}} = \bar{\eta}_{t+1} \\
    \Rightarrow \;\; & \sigma^2 = \rho^2_{t+1}\frac{\bar{\eta}_{t+1}\eta_t}{\bar{\eta}_t} > 0
\end{align}
The above equation has root that $\sigma = \rho_{t+1}\sqrt{{\bar{\eta}_{t+1}\eta_t}\big/{\bar{\eta}_t}}$,
This implies that we can find an independent noise term added to $\vy_t$ to obtain $\vy_{t+1}$, establishing that ${\vy_t}_t$ constitutes a \emph{Markovian forward process}.
\begin{align}
    p(\vy_{t+1}|\vy_t) = \mathcal{N}\left(
    \frac{\rho_{t+1}\bar{\eta}_{t+1}}{\rho_{t}\bar{\eta}_{t}}\vy_t, \rho^2_{t+1}{\frac{\bar{\eta}_{t+1}\eta_t}{\bar{\eta}_t}\mathbf{I}} 
    \right)
\end{align}

What's more, $\{\vx_t\}_t$ sequence can be uniquely determined from $\{\vy_t\}_t$ via 
\begin{align}
 \vx_t = 
 \begin{cases}
 \left(\dfrac{\vy_t}{\rho_t} - \dfrac{\vy_{t+1}}{\rho_{t+1}}\right) \dfrac{1-\gamma_t}{\sqrt{\gamma}_t} ,  & \text{if } t < T \\
 \dfrac{\vy_t}{\rho_t}\dfrac{1-\gamma_t}{\sqrt{\gamma}_t} , & \text{if } t = T.
 \end{cases}
\end{align}

%$\vx_t = \frac{\vy_{t}-\mu_t\vy_{t+1}}{1-\mu_t}$. \jg{redo}

Therefore, the two processes $\{\vx_t\}_t$ and $\{\vy_t\}_t$ has a one-to-one correspondence.

\end{proof}

\section{Implementation Details}
\label{app:implement}

\subsection{Architecture}
\label{app:config}

In this paper, we use encoder-decoder Transformer architecture \citep{transformer} to implement our \modelname{} model. The total number of parameters in our standard model is about 800M. For text-to-image  (T2I) generation, we employ pretrained Flan-T5-XL~\citep{raffel2020exploring,t5xl} as the textual encoder in all experiments. During training, the encoder is frozen and only the decoder is trained. 
For class-conditioned (C2I) generation, the encoder is simply a embedding layer, mapping a class label into a fixed length vector. Since we do not use a text encoder for cross-attention, we pass context information through Adaptive LayerNorm. To maintain parity in the total number of parameters with text-to-image models, we set the hidden size to $1152$.
Below is the default configurations of \modelname{}s. 

\begin{verbatim}
model config for DARTs:
    patch_size=2
    hidden_size=1280 (T2I) or 1152 (C2I)
    num_layers=28
    num_channels_per_head=64
    use_swiglu_ffn=True
    use_rope=True
    rope_axes_dim=[16,24,24]
    use_per_head_rmsnorm=True
    use_adaln=False (T2I) or True (C2I)
    lm_feature_projected_channels=2048
\end{verbatim}

% \begin{verbatim}
% model config:
%     patch_size=2
%     hidden_size=1152 (for 
%     num_layers=28
%     num_channels_per_head=64
%     use_swiglu_ffn=True
%     use_rope=True
%     rope_axes_dim=[16,24,24]
%     use_per_head_rmsnorm=True
%     use_adaln=True
%     lm_feature_projected_channels=2048
% \end{verbatim}

\cref{tb:config} lists the configurations of different model sizes used in the scalability experiments of \S 4.2. For \modelname{}-FM, we implement the flow network as three MLPs (FFNs) with additional adaptive LayerNorm for modulation. Unlike standard FFN blocks in Transformers, the hidden size in our implementation remains unchanged, matching the hidden dimension of the main \modelname{} blocks.

\begin{table}[tbh!]
\begin{center}
\caption{Configurations of \modelname{} of different sizes.}
\label{tb:config}
\small
\begin{tabular}{lcccc}
    \toprule
    % \midrule
    Model  & \# Layers & Hidden size & \# Heads & \# Params \\
    \midrule
    \modelname{}-S & 12 & 384 & 6 & 48M \\
    \modelname{}-B & 12 & 768 & 12 & 141M \\
    \modelname{}-L & 24 & 1024 & 16 & 464M \\
    \modelname{}-XL & 28 & 1280 & 20 & 812M \\
    \bottomrule
  \end{tabular}
\end{center}
\end{table}

In Matryoshka-\modelname{} upsample tuning, we follow the default setting and set the patch size as 2 for $512 \times 512$ images, which end up containing 1024 tokens each. In total, the model has 8192 tokens in total, including 4096 tokens for $256 \times 256$ images with 16 denoising steps and 4096 tokens for $512 \times 512$ images with 4 denoising steps. We also applies rotary positional embedding (RoPE)~\citep{su2024roformer} to embed the information of token position. We find it critical to spatially align the positional embedding of high resolution images with low resolution ones. In particular, we define the rotary matrix for feature $x^{(i,j)}$ at position $(i,j)$ in a $256 \times 256$ image as $\mathbf{R}_{\Theta}^{(i,j)}$. %shown in \cref{eq:rope_2d}, 
%where $d$ is the dimension of the feature and $\Theta = \{ \theta_1, \dots, \theta_{d/2} \}$ are pre-defined frequencies for different feature dimensions. 
For feature $x^{(i',j')}$ at position $(i',j')$ in the corresponding high resolution image with upsample ratio $r$, the rotary matrix is given $\mathbf{R}_{\Theta}^{(i'/r,j'/r)}$. In our case, the ratio is 2, which halves the position values on $512 \times 512$ images. %to align with the lower resolution image. 

% Namely, the position for $512 \times 512$ images are rescaled to have the same maximal value as $256 \times 256$ images. 
% \begin{equation}
%     % f_{\{q.k\}}(x_{(i,j)}, \Theta) = \mathbf{R}_{\Theta,m}^d \mathbf{W}_{\{q,k\}} x_m, \\
%     \mathbf{R}_{\Theta}^{(i,j)} = \begin{pmatrix}
%     \cos i\theta_1 & -\sin i\theta_1 & 0 & 0 & \cdots & 0 & 0 \\
%     \sin i\theta_1 & \cos i\theta_1  & 0 & 0 & \cdots & 0 & 0 \\
%     0 & 0 & \cos j\theta_2 & -\sin j\theta_2 & \cdots & 0 & 0 \\
%     0 & 0 & \sin j\theta_2 &  \cos j\theta_2 & \cdots & 0 & 0 \\
%     \vdots & \vdots & \vdots & \vdots & \ddots & \vdots & \vdots \\
%     0 & 0 & 0 & 0 & \cdots & \cos j\theta_{d/2} & -\sin j\theta_{d/2} \\
%     0 & 0 & 0 & 0 & \cdots & \sin j\theta_{d/2} &  \cos j\theta_{d/2} 
%     \end{pmatrix}
% \label{eq:rope_2d}
% \end{equation}

\subsection{Training}

In all the experiments, we share the following training configuration for our proposed \modelname{}. 

\begin{verbatim}
default training config:
    batch_size=128
    optimizer='AdamW'
    adam_beta1=0.9
    adam_beta2=0.95
    adam_eps=1e-8
    learning_rate=3e-4
    warmup_steps=10_000
    weight_decay=0.01
    gradient_clip_norm=2.0
    ema_decay=0.9999
    mixed_precision_training=bf16
\end{verbatim}

\subsection{Parameterization of Diffusion Model}

In our implementation, we employ v-prediction~\citep{salimans2022progressive} for improved performance. Namely, \modelname{} is trained to output $\vv_t = \frac{\alpha_t \vx_t - \vx}{\sigma_t}$. The prediction of the model $\tilde{\vv}_t$ is mapped to the clean image through $\tilde{\vx} = \alpha_t \vx_t - \sigma_t \tilde{\vv}_t$. The loss is then computed between $\vx$ and $\tilde{\vx}$, and the loss weighting is also applied here. 

\section{Comparison of DART Variants}

We here further clarify the differences and connections between the variants of DART. \Cref{tb:dart_variant} lists the major comparison between DART, DART-AR, and DART-FM. Conceptually, DART predicts the denoised value and adds independent noise to acquire a less noisy image at each step. It conducts this denoising process autoregressively until the clean image is generated. DART-AR applies a token-wise autoregression instead of block-wise autoregressive in vanilla DART, which conduct denoising generation in a more fine-grained granularity. DART-FM, on the other hand, keeps the block-wise autoregression while introduces an additional flow network to conduct flow-mating-based refinement for generated tokens. Both DART-AR and DART-FM improve the performance over vanilla DART. In general, DART-FM demonstrates a better tradeoff between generation quality and inference efficiency. \cref{tb:compare_dart} summarizes the difference between \modelname{} variants.

Kaleido-DART is for multimodal text-image generation, which integrates next-token prediction for text and next-denoising prediction for images (proposed in our DART). In image generation, it can seamlessly adapt all three variants of proposed methods: DART, DART-AR, DART-FM. Similarly, Matryoshka-DART, which enables multi-resolution generation, also adapts to all the three DART variants. Since in Matryoshka-DART, one can simply concatenate high-resolution image tokens after the low-resolution ones, which doesn’t affect the denoising modeling in these variants. 

\begin{table}[htb!]
\caption{Comparison of DART variants.}
\label{tb:compare_dart}
\begin{center}
\begin{tabular}{l c c c c c}
\toprule
Model & Attn Mask & \#AR steps & \#FM steps \\
\midrule
DART & Block-wise Causal & 16 & 0 \\
DART-AR & Causal & 4096 & 0 \\
DART-FM & Block-wise Causal & 16 & 1600 \\
\bottomrule
\end{tabular}
\end{center}
\label{tb:dart_variant}
\end{table}

\section{Additional Samples}
\label{app:additional_samples}

\subsection{Step-wise Generative Results}

\begin{figure}[t]
    \centering
    \includegraphics[width=0.98\linewidth]{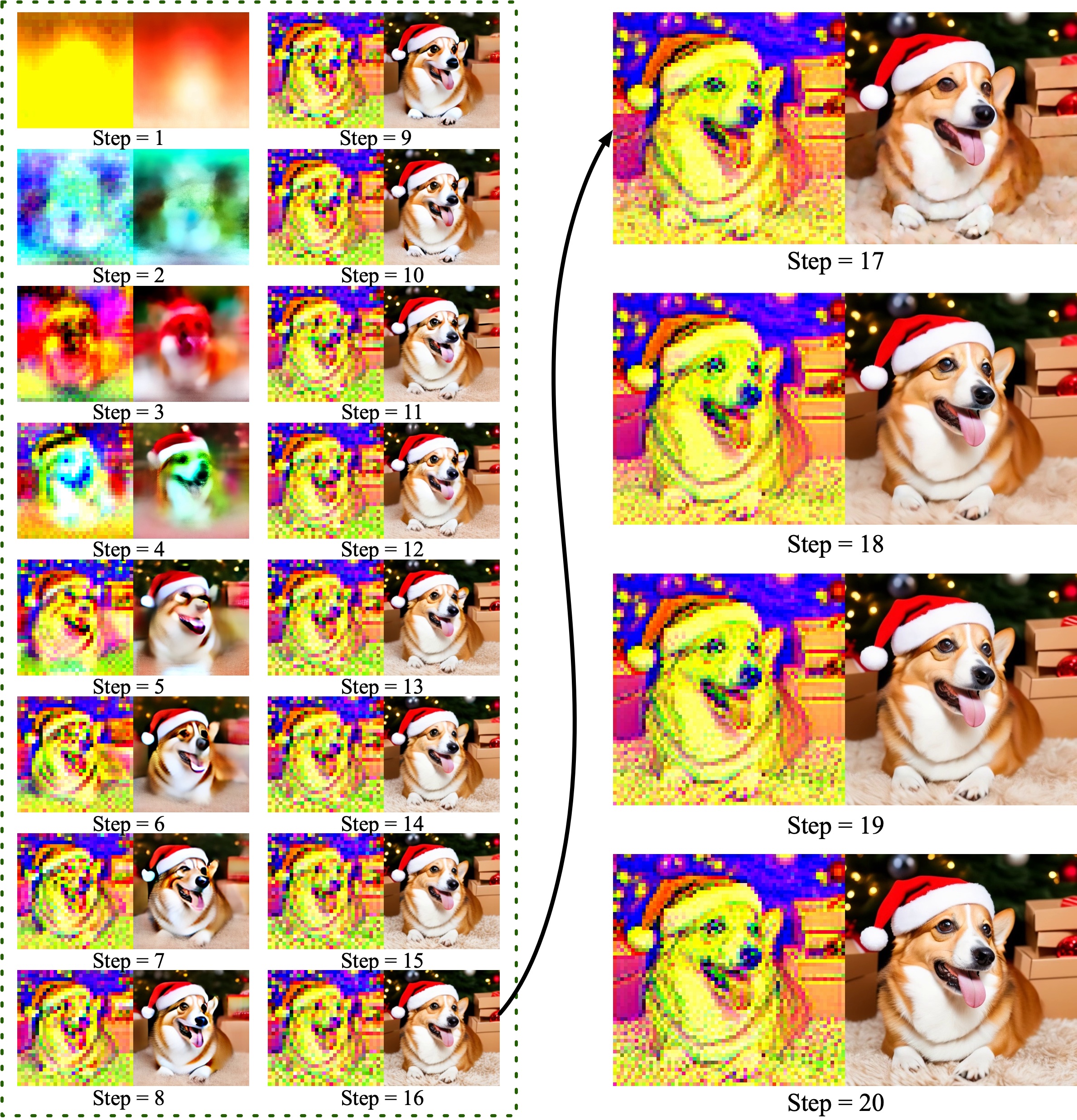}
    \caption{Visualization of the generation process (left: latent maps, right: decoded RGB images) for $256\times 256$ $(T=16)$ and its upsampling to $512\times 512$ $(T=4)$ using Matryoshka-\modelname{}.}
    \label{fig:steps}
\end{figure}

\cref{fig:steps} visualizes the generative process of an image at resolution $256 \times 256$ and its upsampling to $512 \times 512$ by Matryoshka-\modelname{}. \modelname{} first iteratively refines the generative results for 16 denoising steps at resolution $256 \times 256$. It then upsamples images at resolution $512 \times 512$ through iterative denoising as well. Since the generation of high resolution is conditioned on the previous low-resolution samples, it only needs 4 denoising steps at high resolution to generate realistic images.

% \subsection{Comparison of \modelname{} \& \modelname{}-FM}

% We compare the qualitative performance of \modelname{} and \modelname{}-FM using same seeds. It is clearly shown that \modelname{}-FM generates images with higher fidelity and more details when compare with origin \modelname{}, indicating the effectiveness the flow-matching network for improvement. 

\subsection{Additional Samples}

\begin{figure}[t]
    \centering
    \includegraphics[width=0.98\linewidth]{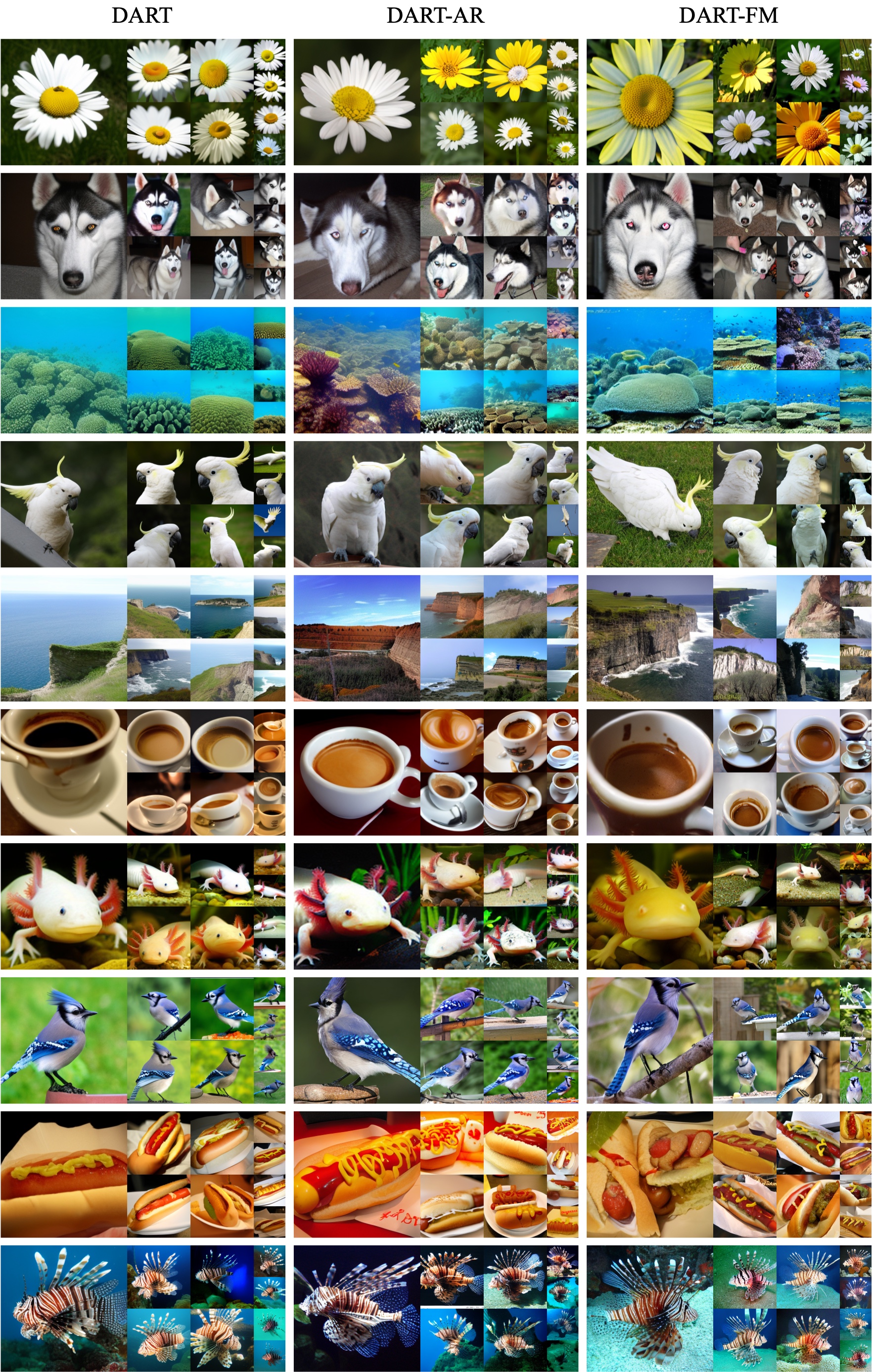}
    \caption{Uncurated samples from \modelname{} varints on ImageNet $256 \times 256$ with for labels of \emph{daisy}, \emph{husky}, \emph{coral reel}, \emph{sulphur-crested cockatoo}, \emph{cliff}, \emph{espresso}, \emph{axolotl}, \emph{jay}, \emph{hotdog}, \emph{lionfish}. }
    \label{fig:additional_imagenet}
\end{figure}

\paragraph{ImageNet}
We here show more generative examples from \modelname{} variants trained on ImageNet $256 \times 256$ in \cref{fig:additional_imagenet}. The proposed two improvement methods, \modelname{}-AR and \modelname{}-FM, generate images with higher fidelity and more details when compared with origin \modelname{}. 

\begin{figure}[t]
    \centering
    \includegraphics[width=\linewidth]{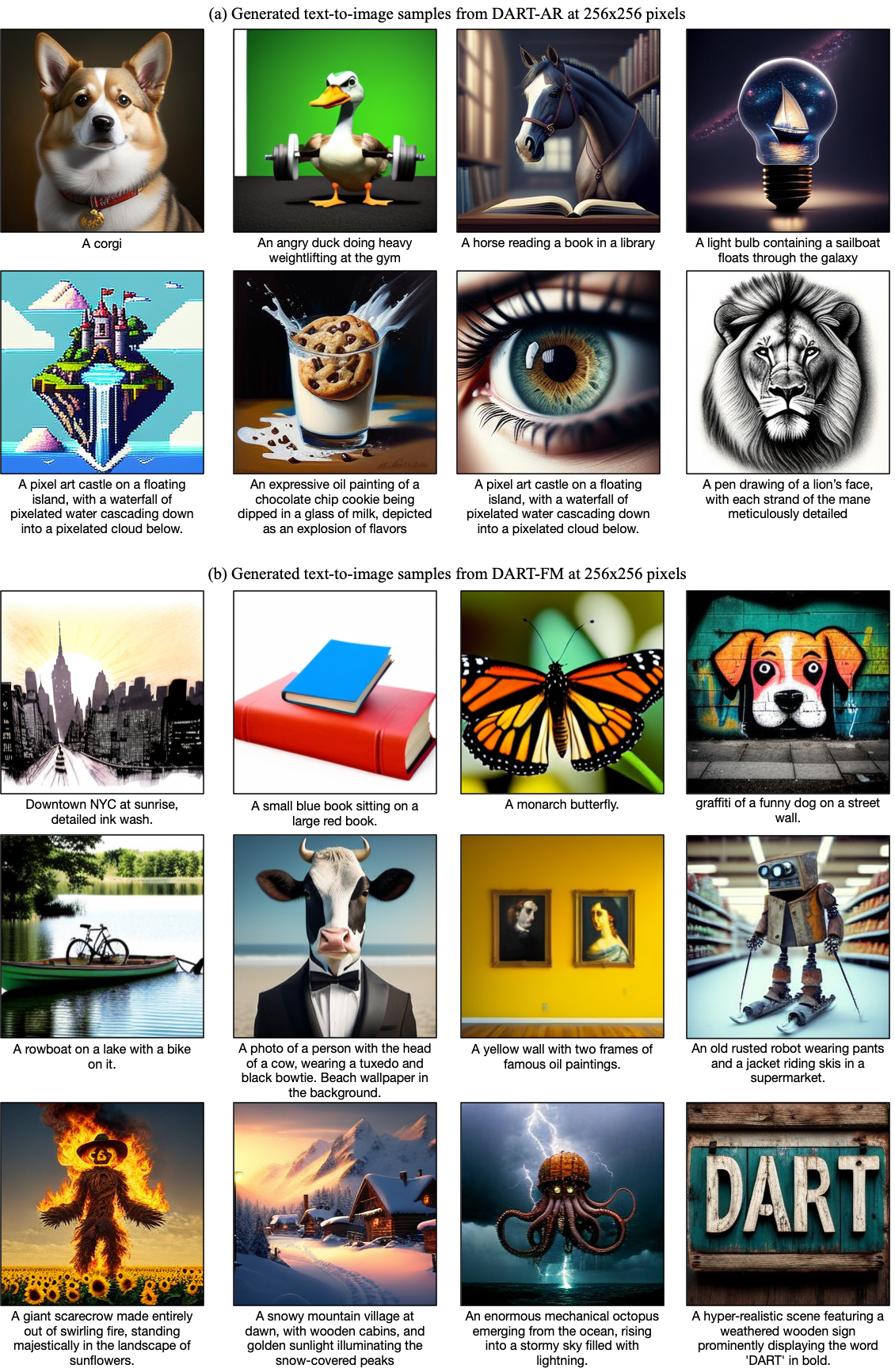}
    \caption{Uncurated samples from \modelname{} varints on text-to-image generation at $256\times 256$ pixels given various captions. }
    \label{fig:additional_t2i_256}
\end{figure}

\begin{figure}[t]
    \centering
    \includegraphics[width=\linewidth]{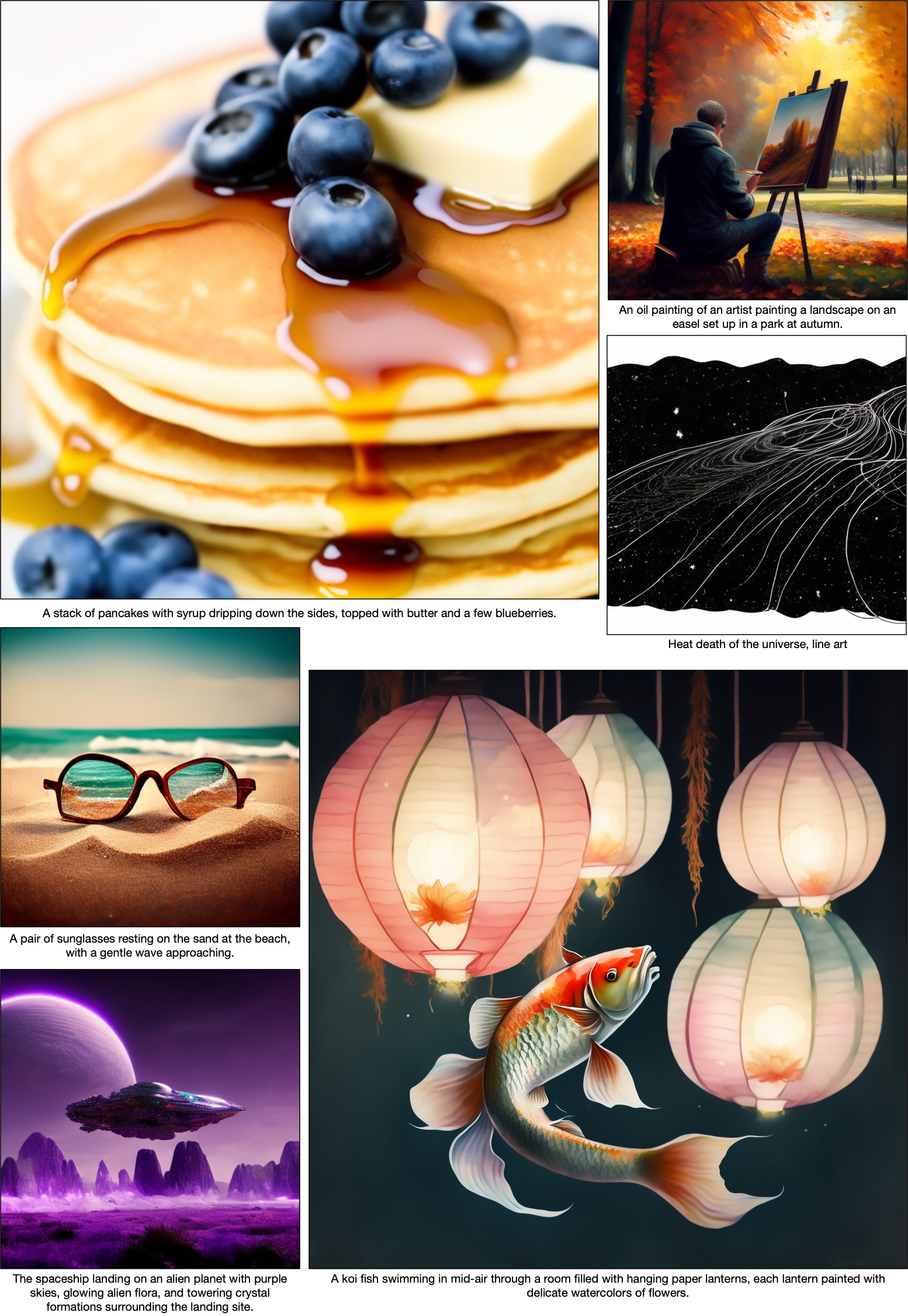}
    \caption{Uncurated samples from \modelname{}-FM with Matryoshka-\modelname{} fine-tuning on text-to-image generation at $512\times 512$ and $1024 \time 1024$ pixels given various captions. }
    \label{fig:additional_t2i_512}
\end{figure}

\begin{figure}[t]
    \centering
    \includegraphics[width=\linewidth]{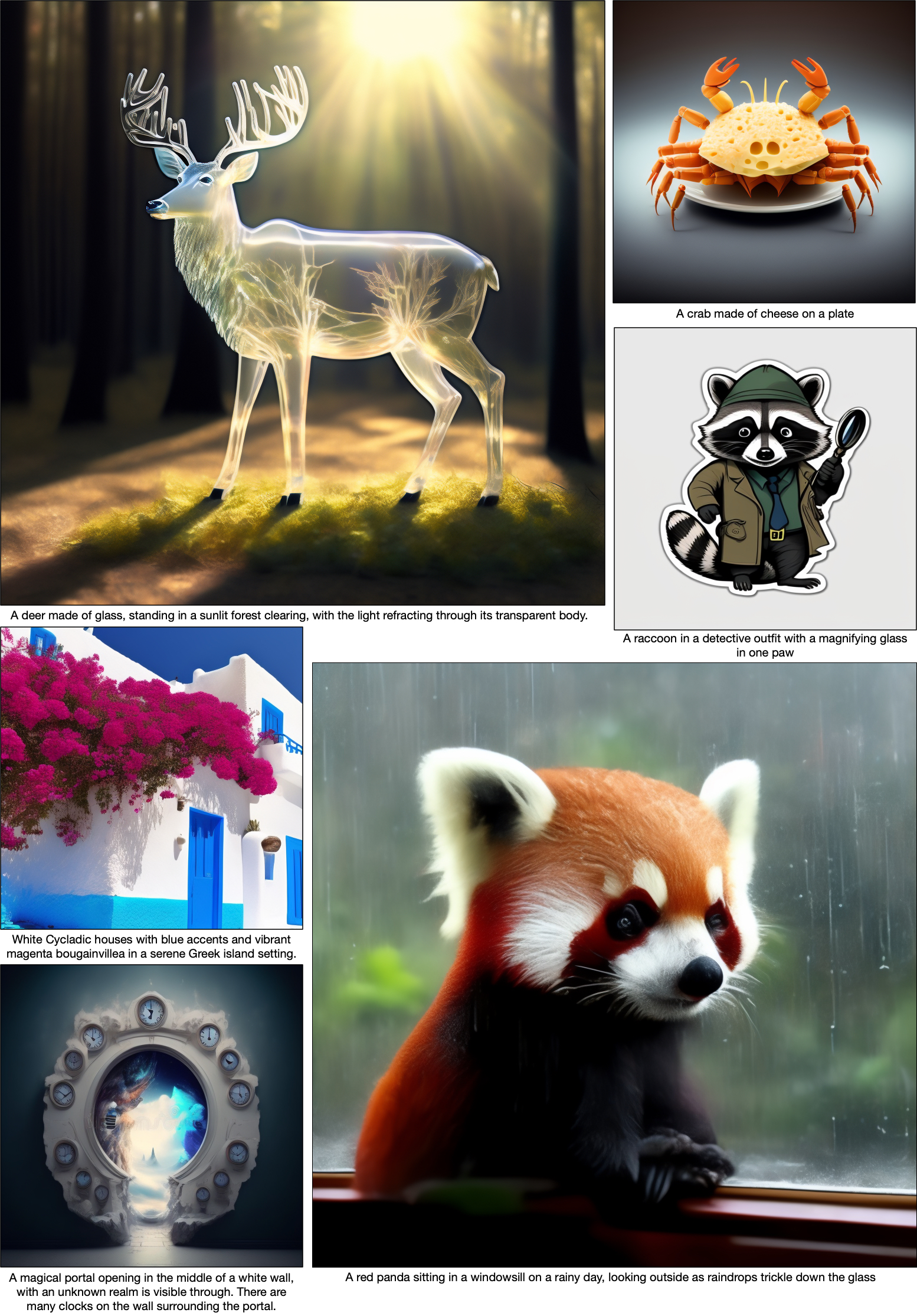}
    \caption{Uncurated samples from \modelname{}-FM with Matryoshka-\modelname{} fine-tuning on text-to-image generation at $512\times 512$ and $1024 \time 1024$ pixels given various captions. }
    \label{fig:additional_t2i_5122}
\end{figure}

\paragraph{Text-to-Image}
We here show more text-to-image generative examples from \modelname{}-AR and \modelname{}-FM at resolution $256 \times 256$ in \cref{fig:additional_t2i_256}. We also show examples at resolutions $512 \times 512$  and $1024 \time 1024$ from Matryoshka-\modelname{} finetuned from \modelname{}-FM at $256 \times 256$ in \cref{fig:additional_t2i_512,fig:additional_t2i_5122}. 

% \subsection{Additional Multimodal Generation Samples (Optional?)}

% We here show more text-image co-generation examples from Kaleido-\modelname{}.

\end{document}